\documentclass[twocolumn,letterpaper]{IEEEAerospaceCLS}  % only supports two-column, letterpaper format

\usepackage[]{graphicx}    % We use this package in this document
\usepackage{amsmath}
\usepackage{amsfonts}
\usepackage{subcaption}
\usepackage{fancyhdr}
\usepackage{booktabs}
\usepackage[colorlinks=true, urlcolor=blue, linkcolor=black, citecolor=black]{hyperref} % ED 2025 added for web links
\newcommand{\ignore}[1]{}  % {} empty inside = %% comment
\begin{document}
\title{Distributed Area Coverage with High Altitude Balloons Using Multi-Agent Reinforcement Learning}

\fancypagestyle{firstpage}{
    \fancyhf{} % Clear all headers and footers
    \fancyfoot[C]{DISTRIBUTION STATEMENT A: Approved for public release, distribution is unlimited.} % Footer in the center
    \renewcommand{\headrulewidth}{0pt}
}
\pagestyle{plain}

%\maketitle
%\thispagestyle{firstpage}
%\pagestyle{plain}

\author{%
Adam Haroon\\ %$^{*}$\\ 
NREIP Intern\\
U.S. Naval Research Laboratory\\
4555 Overlook Ave. S.W \\
Washington, D.C. 20375 \\
adam.o.haroon.ctr@us.navy.mil
\and 
Tristan Schuler\\
U.S. Naval Research Laboratory\\
4555 Overlook Ave. S.W \\
Washington, D.C. 20375 \\
tristan.k.schuler.civ@us.navy.mil
%%%% IMPORTANT: Use the correct copyright information--IEEE, Crown, or U.S. government. %%%%%
\thanks{U.S. Government work not protected by U.S. copyright.}
%\thanks{\footnotesize $^{*}$Completed in part during an internship at U.S. Naval Research Laboratory.\\
%979-8-3315-7360-7/26/$\$31.00$ \copyright2026 IEEE}              % This creates the copyright info that is the correct 2026 data.
%\thanks{{U.S. Government work not protected by U.S. copyright}}         % Use this copyright notice only if you are employed by the U.S. Government.
%\thanks{{979-8-3315-7360-7/26/$\$31.00$ \copyright2026 Crown}}          % Use this copyright notice only if you are employed by a crown government (e.g., Canada, UK, Australia).
%\thanks{{979-8-3315-7360-7/26/$\$31.00$ \copyright2026 European Union}}    % Use this copyright notice is you are employed by the European Union.
}

\maketitle

\thispagestyle{plain}
\pagestyle{plain}

\maketitle

\fancypagestyle{firstpage}{
    %\fancyhf{} % Clear all headers and footers
    \fancyfoot[C]{DISTRIBUTION STATEMENT A: Approved for public release, distribution is unlimited.
    \\\thepage} % Footer in the center
    \renewcommand{\headrulewidth}{0pt}
}

\maketitle
\thispagestyle{firstpage}
\pagestyle{plain}

\begin{abstract}
High Altitude Balloons (HABs) can leverage stratospheric wind layers for limited horizontal control, enabling applications in reconnaissance, environmental monitoring, and communications networks. Existing multi-agent HAB coordination approaches use deterministic methods like Voronoi partitioning and extremum seeking control for large global constellations, which perform poorly for smaller teams and localized missions. While single-agent HAB control using reinforcement learning has been demonstrated on HABs, coordinated multi-agent reinforcement learning (MARL) has not yet been investigated. This work presents the first systematic application of multi-agent reinforcement learning (MARL) to HAB coordination for distributed area coverage. We extend our previously developed reinforcement learning simulation environment (RLHAB) to support cooperative multi-agent learning, enabling multiple agents to operate simultaneously in realistic atmospheric conditions. We adapt QMIX for HAB area coverage coordination, leveraging Centralized Training with Decentralized Execution to address atmospheric vehicle coordination challenges. Our approach employs specialized observation spaces providing individual state, environmental context, and teammate data, with hierarchical rewards prioritizing coverage while encouraging spatial distribution. We demonstrate that QMIX achieves similar performance to the theoretically optimal geometric deterministic method for distributed area coverage, validating the MARL approach and providing a foundation for more complex autonomous multi-HAB missions where deterministic methods become intractable.
\end{abstract}

\tableofcontents

%%%%%%%%%%%%%%%%%%%%%%%%%%%%%%%%%%%%%%
\section{Introduction and Background}
%%%%%%%%%%%%%%%%%%%%%%%%%%%%%%%%%%%%%%

%\subsection{High Altitude Balloons: Capabilities and Control Mechanisms}

Altitude-controllable High Altitude Balloons (HABs) operate primarily in the stratosphere, typically between 15-25 km altitude, leveraging lighter-than-air buoyancy principles to maintain flight for extended time periods~\cite{abe2009scientific}. Unlike conventional aircraft that rely on thrust for both lift and propulsion, HABs achieve lift through lighter-than-air gases or heated ambient air and accomplish horizontal movement by exploiting stratospheric wind patterns at different altitudes. Altitude-controlled HABs employ various mechanisms to adjust their altitude
%and access different wind layers: Solar High Altitude Balloons with Vents (SHAB-Vs) utilize solar heating of ambient air combined with mechanical venting systems~\cite{schuler2023altitude}, while super-pressure and zero-pressure balloons use internal pressure variation or ballast/venting systems respectively~\cite{abe2009scientific}. These control mechanisms enable HABs
to "ride" favorable winds, providing limited but strategic horizontal control for station-keeping and trajectory-following maneuvers.
%~\cite{bowman2020multihour,schuler2023altitude}. 
The amount of horizontal controllability of HABS and station keeping potential is highly dependent on the wind diversity for particular seasons and geographic areas, where Equatorial Regions have high wind diversity year-round and mid-latitudes have favorable conditions in summer months~\cite{schuler2025winddiversity,brown2025stratospheric}. 

%The stratosphere presents unique advantages, including minimal weather interference, predictable wind patterns, and reduced air traffic. 

The stratosphere presents unique challenges including highly complex dynamic flow fields, limited observational wind data leading to forecast uncertainty, and cold operating temperatures.
Successful HAB navigation requires a sophisticated understanding of atmospheric dynamics, as wind velocity and direction vary dramatically across altitude levels, time, and geography. This vertical wind diversity creates opportunities for strategic positioning while presenting significant challenges for autonomous control systems operating in this complex four-dimensional flow field environment.

\subsection{Reinforcement Learning in Aerospace Applications}

Reinforcement Learning (RL) has emerged as a powerful approach for autonomous control in aerospace applications, particularly for systems operating in complex, dynamic environments with partial observability. Single-agent RL has demonstrated recent success across various unmanned aerial vehicle (UAV) domains, including fixed-wing aircraft path planning~\cite{koch2019reinforcement}, rotorcraft control in turbulent conditions~\cite{patino2023learning}, and satellite attitude control~\cite{cai2024reinforcement}.

For atmospheric vehicles specifically, RL has shown promise in addressing the inherent challenges of uncertain wind conditions and forecast inaccuracies. Traditional model-based control approaches struggle with the high dimensionality and non-stationary nature of atmospheric flow fields, making learning-based methods particularly attractive, as first demonstrated by Google Loon~\cite{bellemare2020autonomous}. Deep Q-Networks (DQN) and policy gradient methods have also been successfully applied to various atmospheric navigation problems, demonstrating the ability to learn robust policies that handle forecast uncertainties and dynamic environmental conditions~\cite{gannetti2023navigation,schuler2025seasonal}.

Multi-Agent Reinforcement Learning (MARL) has gained traction in aerospace for coordinating multiple UAVs. Applications include formation flying~\cite{xing2024multi}, distributed surveillance~\cite{yun2022cooperative}, and aerial robotic swarms~\cite{tang2025enhanced}. Among MARL algorithms, QMIX has shown particular promise for cooperative coordination tasks, learning joint action-values as monotonic combinations of individual agent values while enabling tractable policy extraction in cooperative settings~\cite{rashid2020monotonic}. These systems typically involve agents with similar capabilities operating in shared airspace, requiring coordination to avoid conflicts while achieving collective objectives.

\subsection{Multi-Agent Coordination for HABs}

Despite the growing body of research in both single-agent HAB control and other multi-agent aerospace systems, multi-agent coordination with high altitude balloons remains underexplored. Deterministic multi-agent coordination approaches, such as Voronoi partitioning and extremum seeking control, have been proposed to achieve a global constellation with thousands of high-altitude balloons~\cite{du2022dynamic,vandermeulen2017distributed}. These approaches typically do not account for forecast uncertainty, learning capabilities, or the dynamic adaptation requirements essential for effective small-team coordination in uncertain atmospheric environments. Other multi-agent HAB research has focused on individual parallelized station-keeping, treating each balloon as a fully independent agent with its own objectives~\cite{gannetti2023navigation,xu2022station}.

The unique characteristics of HAB flight, including limited direct control authority, dependence on atmospheric conditions, and operational constraints over vast geographical areas, create distinct challenges that existing MARL frameworks have not addressed. Applications such as distributed sensing networks, regional communication coverage, and efficient high-fidelity weather data collection inherently require multiple HABs to operate as coordinated teams rather than independent agents. 
%To address this critical gap in multi-agent HAB coordination, we first established a robust single-agent HAB control as the foundation for our approach.

\subsection{Foundation: Single-Agent HAB Station-Keeping}

Our previous work established the feasibility of autonomous vented solar high altitude balloon (SHAB-V) station-keeping using deep reinforcement learning~\cite{schuler2025seasonal}. Furthermore, this research developed the RLHAB simulation environment, which provides realistic atmospheric modeling using ERA5 reanalysis data and synthetic wind field generation derived from radiosonde observations. The RLHAB framework introduced several key innovations, including synthetic ground truth winds derived from historical radiosonde data, modular weather forecast integration, and a modular structure for streamlined training, evaluating, and extendability of different types of balloon dynamics and/or state-of-the-art reinforcement learning algorithms.

The single-agent framework successfully demonstrated that DQN agents could learn effective station-keeping policies for SHAB-V platforms, achieving time-within-region performance of approximately 50\% across various seasonal conditions, with performance varying significantly by season and geographic location. The single-agent approach employed a discrete action space consisting of altitude control commands (ascend, maintain, descend) and utilized a piecewise reward function based on distance to target regions. Agents learned to exploit vertical wind diversity by changing altitude to access more favorable wind conditions, effectively using the three-dimensional atmospheric environment for horizontal positioning control.
%However, the original single-agent framework was fundamentally limited to individual objectives. Each balloon operates independently with its own assigned coverage region, receiving rewards based solely on its own individual performance. 
However, this initial framework did not address scenarios where mission success depends on coordination between multiple HABs, spatial distribution optimization, or dynamic reallocation of coverage responsibilities based on changing atmospheric conditions.

\subsection{Distributed Area Coverage with HABs}

Many practical HAB applications require capabilities that extend beyond individual station-keeping to coordinated area coverage missions. \textbf{Distributed area coverage} refers to the coordinated positioning of multiple HAB agents to maximize spatial coverage over a target geographical region while maintaining optimal agent distribution to avoid redundancy, clustering, and coverage gaps. This represents a fundamental shift from individual agent objectives to team-based performance optimization.
%Emergency response scenarios exemplify the importance of distributed coverage capabilities. During natural disasters, HABs can provide critical communication relay services over affected regions, but effective coverage requires strategic positioning of multiple agents to ensure continuous connectivity across the disaster area. Similarly, environmental monitoring missions benefit from coordinated sensor networks that optimize spatial sampling while adapting to changing atmospheric conditions that can temporarily render certain regions inaccessible.
Multi-agent coordination offers several key advantages for HAB area coverage missions: improved spatial distribution and persistent coverage, adaptive response to changing wind conditions through distributed decision-making, and resource efficiency through minimized redundant coverage.

The distributed coverage problem introduces several challenges that single-agent approaches cannot address, including spatial interdependence where optimal positioning for one agent depends critically on the positions of all other agents, dynamic reallocation requirements when atmospheric conditions change, and resource optimization for efficient coverage with finite agents over large geographical areas.
The unique constraints of HAB operations, including limited control authority, dependence on atmospheric wind patterns, and potential communication limitations over vast operational distances, compound these challenges. Traditional multi-agent coordination approaches developed for powered aircraft may not translate directly to the atmospheric vehicle domain, necessitating specialized methods that account for HAB-specific operational characteristics.

%\subsection{Paper Structure}

%The remainder of this paper is organized as follows. Section 2 presents the preliminaries for reinforcement learning and the distinction between single-agent and multi-agent RL. Section 3 outlines the theoretical methodology, comparing MARL paradigms and justifying our selection of the QMIX algorithm for atmospheric vehicle coordination. Section 4 details our experimental setup, including the multi-agent simulation environment, observation space design, and cooperative reward structures. Section 5 presents our results and analysis, demonstrating the effectiveness of cooperative MARL for distributed HAB coverage tasks. Section 6 concludes with a discussion of the implications of this work and future research directions.

%%%%%%%%%%%%%%%%%%%%%%%%%%%%%%%%%%%%%%
\section{Preliminaries for HAB MARL}
%%%%%%%%%%%%%%%%%%%%%%%%%%%%%%%%%%%%%%

\subsection{Single-Agent vs. Multi-Agent Reinforcement Learning}

Traditional single-agent reinforcement learning addresses sequential decision-making problems in which a single agent learns to maximize cumulative reward through interaction with an environment. The agent observes states $s_t$, takes actions $a_t$, and receives rewards $r_t$, with the objective of learning an optimal policy $\pi^*(s)$ that maximizes the expected discounted return:

\begin{equation}
J(\pi) = \mathbb{E}_{\pi}\left[\sum_{t=0}^{\infty} \gamma^t r_t\right]
\end{equation}

where $\gamma \in [0,1]$ is the discount factor. In our previous work on single-agent HAB station-keeping~\cite{schuler2025seasonal}, this framework was sufficient since each balloon operated independently with the goal of station-keeping within its assigned target region and received rewards based solely on its own performance.

Multi-Agent Reinforcement Learning (MARL) extends this paradigm to environments with multiple learning agents, leading to significantly more complex dynamics~\cite{huh2023multi}. The presence of multiple agents fundamentally alters the environment from the perspective of any individual agent, as other agents' policies are simultaneously evolving during training. This creates a non-stationary environment where the Markov property may be violated from each agent's local perspective~\cite{hernandez2017survey}.

For $n$ agents, where $n$ denotes the number of agents in the system and $|\mathcal{A}|$ represents the size of an individual agent's action space, the joint action space grows exponentially as $|\mathcal{A}|^n$, leading to the curse of dimensionality~\cite{gronauer2022multi}. Additionally, the credit assignment problem becomes critical: When agents receive team rewards, determining each agent's individual contribution to the overall outcome becomes non-trivial~\cite{huh2023multi}. These challenges necessitate specialized algorithms designed specifically for multi-agent coordination.

\subsection{Multi-Agent Reinforcement Learning Paradigms}

MARL approaches can be categorized into three primary paradigms based on the nature of agent interactions and reward structures~\cite{zhang2021multi,huh2023multi}:

\subsubsection{Independent Learning}

Independent learning treats each agent as operating in isolation, applying single-agent RL algorithms while ignoring the presence of other agents~\cite{tan1993multi}. Each agent $i$ learns its own policy $\pi_i$ based on its individual reward signal $r_i^t$ by treating other agents as part of the environment dynamics:

\begin{equation}
J_i(\pi_i) = \mathbb{E}_{\pi_i}\left[\sum_{t=0}^{\infty} \gamma^t r_i^t\right]
\end{equation}

While computationally simple and highly scalable, this approach can be effective when agents operate fully independently towards their own objective without any influence on the behavior of other agents or the environment itself. However, when attempting to model agent interaction, communication, or coordination, independent learning suffers from non-stationarity issues as other agents' policies evolve during training, potentially leading to unstable learning and suboptimal coordination~\cite{hernandez2017survey}.

\subsubsection{Competitive Learning}

Competitive MARL addresses zero-sum or adversarial scenarios where agents have conflicting objectives~\cite{littman1994markov}. In competitive settings, each agent seeks to maximize its individual reward where the environment structure creates inherently conflicting objectives between agents, often through zero-sum reward structures~\cite{yang2020overview}. Although atmospheric vehicles could theoretically compete for favorable wind resources, this is counterproductive for distributed area coverage, where coordination among all agents is required to maximize performance. 

\subsubsection{Cooperative Learning}

Cooperative MARL addresses scenarios where agents share common objectives or complementary goals~\cite{yuan2023survey}. Agents aim to maximize team performance rather than individual rewards:

\begin{equation}
J_{team}(\pi_1, ..., \pi_n) = \mathbb{E}_{(\pi_1, ..., \pi_n)}\left[\sum_{t=0}^{\infty} \gamma^t R_{team}^t\right]
\end{equation}

where $R_{team}^t$ represents the shared team reward. This paradigm is essential for distributed area coverage, where individual agent success is meaningless without effective team coordination. 
%Having established the cooperative learning paradigm, we now formalize the distributed coverage problem and detail our QMIX-based approach.

\section{Distributed Area Coverage: QMIX}

%For this work, distributed area coverage refers to the coordinated positioning of multiple HAB agents to maximize spatial coverage over a target geographic region while maintaining optimal agent distribution to avoid redundancy, clustering, and gaps. This problem inherently requires cooperative behavior for several fundamental reasons:

Distributed area coverage inherently requires coordinated cooperative behavior among HAB agents for several fundamental reasons:

\textbf{Shared Objective}: All agents share the common goal of maximizing coverage of the target area region, making team success the primary metric rather than individual performance.

\textbf{Spatial Interdependence}: Agent positioning decisions are inherently interdependent; Optimal placement for one agent depends critically on the positions of all other agents to avoid redundant coverage, gaps, or clustering.

\textbf{Resource Constraints}: With finite agents and coverage requirements, coordination is essential to efficiently allocate limited resources across a large area.

Independent learning fails for this problem because agents cannot assess how their actions affect overall team performance without explicit coordination mechanisms, which in turn introduces non-stationarity. Cooperative learning approaches are therefore essential, with value decomposition methods being particularly well-suited due to their ability to handle shared rewards while maintaining individual decision-making capabilities~\cite{sunehag2017value}.

\subsection{QMIX Algorithm Selection and Overview}

Among cooperative MARL algorithms, we selected QMIX for its specific advantages in atmospheric vehicle coordination tasks~\cite{rashid2020monotonic}. In distributed HAB coverage, agents must coordinate through shared atmospheric resources (wind fields) while making individual altitude decisions that affect the team-wide spatial distribution. QMIX's value decomposition approach enables agents to learn how their individual altitude choices contribute to collective coverage objectives while operating in the same dynamic wind environment.

QMIX offers several advantages over alternative approaches. Compared to policy-gradient methods like MADDPG~\cite{lowe2017multi}, QMIX's value-based formulation provides more stable learning in our discrete altitude control space, which naturally aligns with practical HAB systems that operate through distinct ascent, maintain, and descent modes. While simpler value decomposition methods like VDN~\cite{sunehag2017value} assume additive value functions, QMIX's mixing network can capture complex interactions between agents navigating shared wind resources. Unlike methods requiring separate critics for each agent (such as COMA~\cite{foerster2018counterfactual}), QMIX employs a single mixing network that scales more efficiently with the size of the fleet while effectively addressing the problem of credit assignment.

\begin{figure*}
\centering
\includegraphics[width=\textwidth]{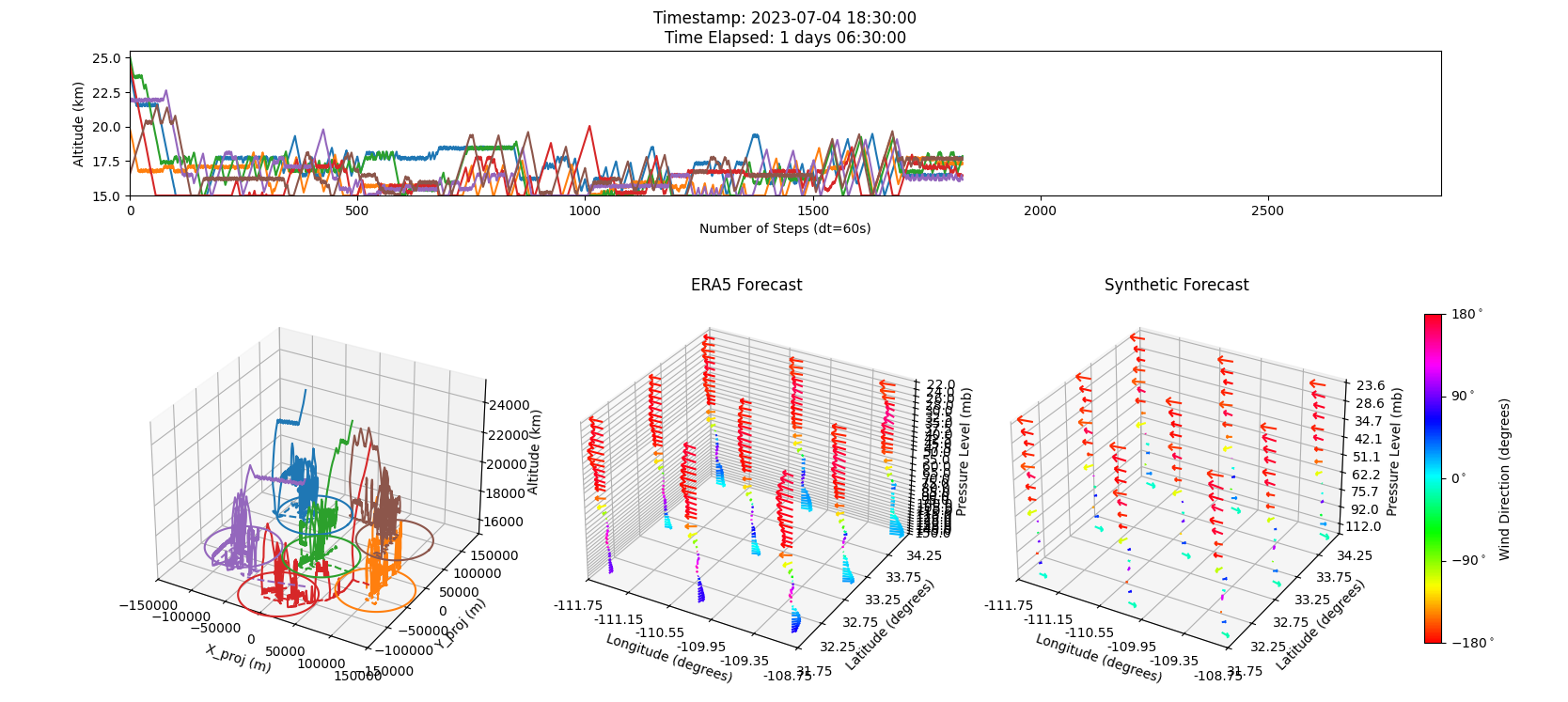}
\caption{RLHAB v.2 simulation environment extended to support multi-agent dynamics and control with different forecast and ground truth wind flow fields.}
\label{fig:RLHAB-sim}
\end{figure*}

\subsubsection{Centralized Training with Decentralized Execution}
The CTDE framework enables agents to leverage global information during training while operating independently during execution~\cite{amato2024introduction}. During training, agents access complete atmospheric conditions and all agent positions, enabling sophisticated coordination strategies. During execution, agents operate using only local observations and trained individual Q-networks, eliminating real-time coordination requirements while maintaining learned cooperative behaviors.

\subsubsection{Individual-Global-Max Principle}
QMIX ensures that the optimal joint action according to the centralized Q-function $Q_{tot}$ corresponds to each agent selecting the action with the highest individual Q-value~\cite{yuan2023survey}. This Individual-Global-Max (IGM) condition is expressed as:
\begin{equation}
\arg \max_{\mathbf{a}} Q_{tot}(\boldsymbol{\tau}, \mathbf{a}, s) = \begin{pmatrix}
\arg \max_{a^1} Q_1(\tau^1, a^1) \\
\vdots \\
\arg \max_{a^n} Q_n(\tau^n, a^n)
\end{pmatrix}
\end{equation}
where $\boldsymbol{\tau} = (\tau^1, ..., \tau^n)$ represents the joint action-observation history, $\mathbf{a} = (a^1, ..., a^n)$ is the joint action, and $s$ is the global state.

\subsubsection{Mixing Network Architecture}
QMIX uses a mixing network that combines the Q-values of individual agents to produce the total Q-value of the team $Q_{tot}$~\cite{rashid2020monotonic}. The architecture uses hypernetworks to generate weights from the global state while ensuring monotonicity: $\frac{\partial Q_{tot}}{\partial Q_i} \geq 0, \forall i$. This constraint ensures that improving any individual agent's performance cannot decrease total team value, preventing conflicting objectives during training.

\subsubsection{QMIX Application to HAB Coverage}
For distributed HAB coverage, QMIX provides several key advantages: the mixing network directly optimizes team-level coverage metrics while maintaining individual decision-making capabilities; the CTDE framework enables stable learning with shared rewards while allowing trained agents to operate independently using only local observations; and the monotonicity constraint ensures that improvements in individual agent performance contribute positively to team objectives, crucial for HAB missions where communication may be intermittent across vast operational distances. Having established the theoretical foundation for applying QMIX to distributed HAB coverage, we now detail our experimental implementation. The following section describes how we adapt the RLHAB simulation environment for multi-agent scenarios, design observation and reward structures that enable effective coordination learning, and establish evaluation criteria to assess the effectiveness of our cooperative MARL approach.

%%%%%%%%%%%%%%%%%%%%%%%%%%%%%%%%%%%%%%
\section{RLHAB-MARL Experimental Setup}
%%%%%%%%%%%%%%%%%%%%%%%%%%%%%%%%%%%%%%

\begin{figure*}[btp] % spans across 2 columns
    \centering
    
    % Each image takes ~0.19 of textwidth (5 across)
    \begin{subfigure}{0.19\textwidth}
        \centering
        \includegraphics[width=\linewidth]{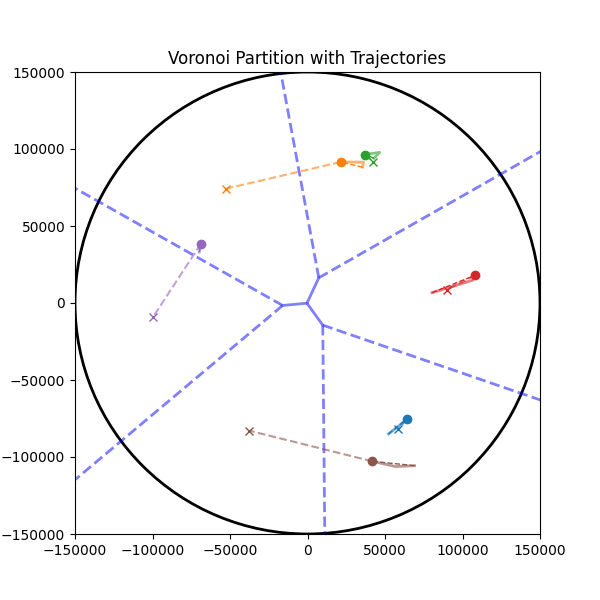}
        %\caption{A}
        \label{fig:5a}
    \end{subfigure}
    \hfill
    \begin{subfigure}{0.19\textwidth}
        \centering
        \includegraphics[width=\linewidth]{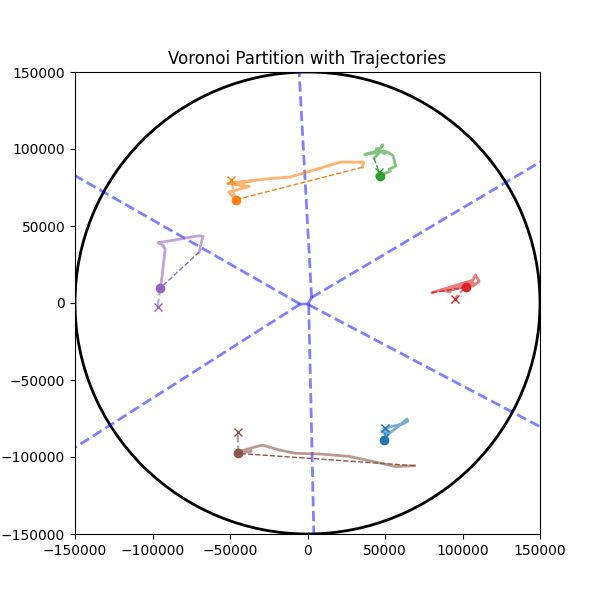}
        %\caption{B}
        \label{fig:5b}
    \end{subfigure}
    \hfill
    \begin{subfigure}{0.19\textwidth}
        \centering
        \includegraphics[width=\linewidth]{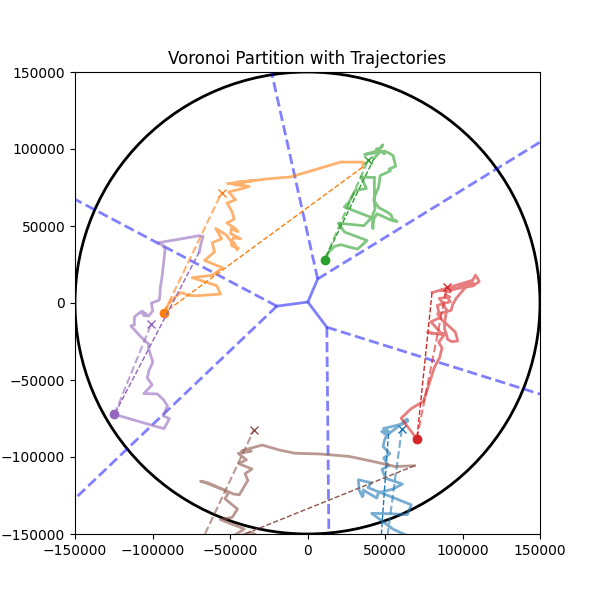}
        %\caption{C}
        \label{fig:5c}
    \end{subfigure}
    \hfill
    \begin{subfigure}{0.19\textwidth}
        \centering
        \includegraphics[width=\linewidth]{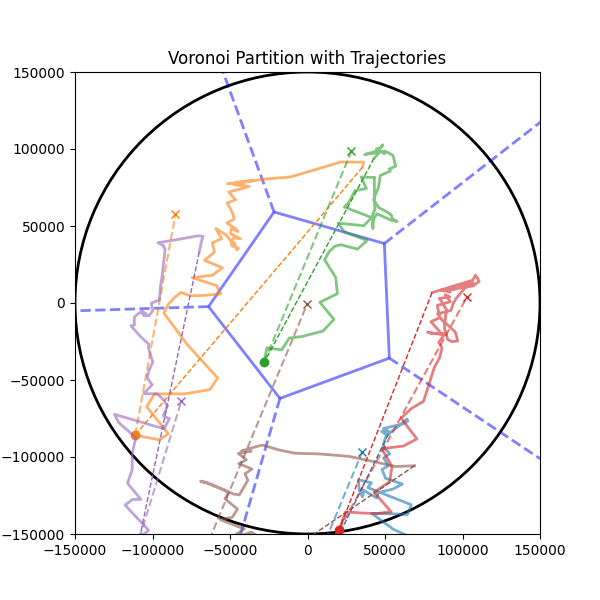}
        %\caption{D}
        \label{fig:5d}
    \end{subfigure}
    \hfill
    \begin{subfigure}{0.19\textwidth}
        \centering
        \includegraphics[width=\linewidth]{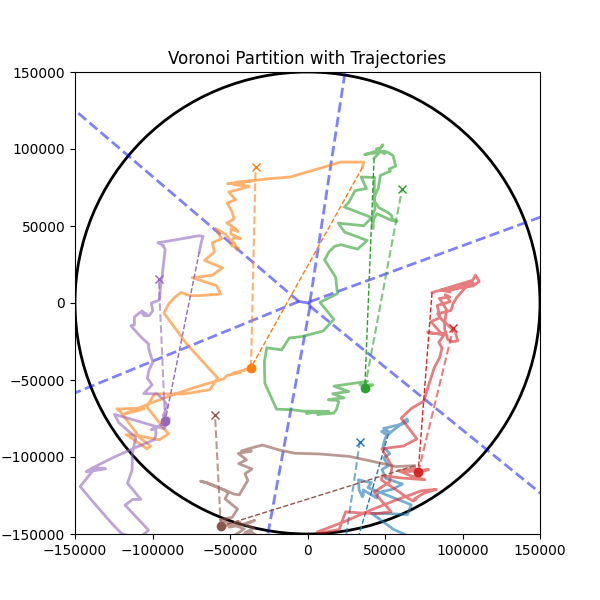}
        %\caption{E}
        \label{fig:5e}
    \end{subfigure}
    
    \caption{Voronoi partitions, centroid waypoints, and HAB trajectories changing over time.}
    \label{fig:baseline_time}
\end{figure*}

\subsection{RLHAB Multi-Agent Simulation Environment}

Our experimental framework builds upon the RLHAB simulation environment introduced in our previous work~\cite{schuler2025seasonal}, extending it to support cooperative multi-agent reinforcement learning for distributed area coverage missions. The original RLHAB framework provided realistic single-agent HAB station-keeping capabilities using ERA5 reanalysis data and synthetic wind fields for ground truth. For this work, we extended the environment to support multiple simultaneous agents operating under shared atmospheric conditions.  The multi-agent RLHAB environment maintains the core atmospheric modeling fidelity of the original framework, including realistic, uncertain flow fields generated by aggregating radiosonde profiles and using the European Centre for Medium-Range Weather Forecasts (ECMWF) Complete ERA5 Reanalysis as the observable forecast. The simulation operates with a temporal resolution of 1 minute and covers altitude ranges from 15 km to 25 km, corresponding to typical HAB operational regions.  Figure \ref{fig:RLHAB-sim} shows a mid-episode GUI output of the multi-agent simulation environment running the Voronoi baseline controller for 6 agents in favorable station-keeping conditions during a day in July in the Southwestern United States. 
Key modifications for multi-agent scenarios include synchronized agent dynamics, shared atmospheric state propagation, and coordinated observation generation. The environment implements the PettingZoo \texttt{ParallelEnv} interface, enabling streamlined integration with modern MARL algorithms while maintaining computational efficiency. 
%All agents experience identical wind conditions at any given spatiotemporal location, ensuring realistic physics-based interactions through the shared dynamic flow field environment rather than direct agent-to-agent communication.

\subsection{Multi-Agent Environment Design}

For the distributed area coverage task, we configure the environment with $N_{agents}$ HAB agents tasked with cooperatively covering a circular target region. The coverage area has a radius of $R_{coverage} = 3 \times R_{station} = 150$ km, where $R_{station} = 50$ km is the traditional station-keeping radius from our previous work. 
%This configuration represents realistic regional coverage scenarios such as disaster response, environmental monitoring, or communication relay missions.

Agent initialization occurs within the coverage area using randomized positions to ensure diverse starting conditions across episodes. The optimal target agent separation distance is calculated as $d_{target} = R_{coverage}/\sqrt{N_{agents}}$, where $N_{agents}$ is the number of agents. This heuristic approximates the optimal spatial distribution by dividing the circular coverage area equally among agents.

Episode termination occurs based on temporal constraints (maximum episode length) or forecast data availability. Episodes run for 2,880 time steps at 1-minute resolution, corresponding to 48 hours of simulated mission time, providing sufficient duration for meaningful coordination learning while maintaining computational tractability. To evaluate the effectiveness of our cooperative learning approach within this environment, we establish a baseline for comparison.

\begin{figure}[!t] % stays within one column
    \centering
    
    % First row
    \begin{subfigure}{0.48\columnwidth}
        \centering
        \includegraphics[width=\linewidth]{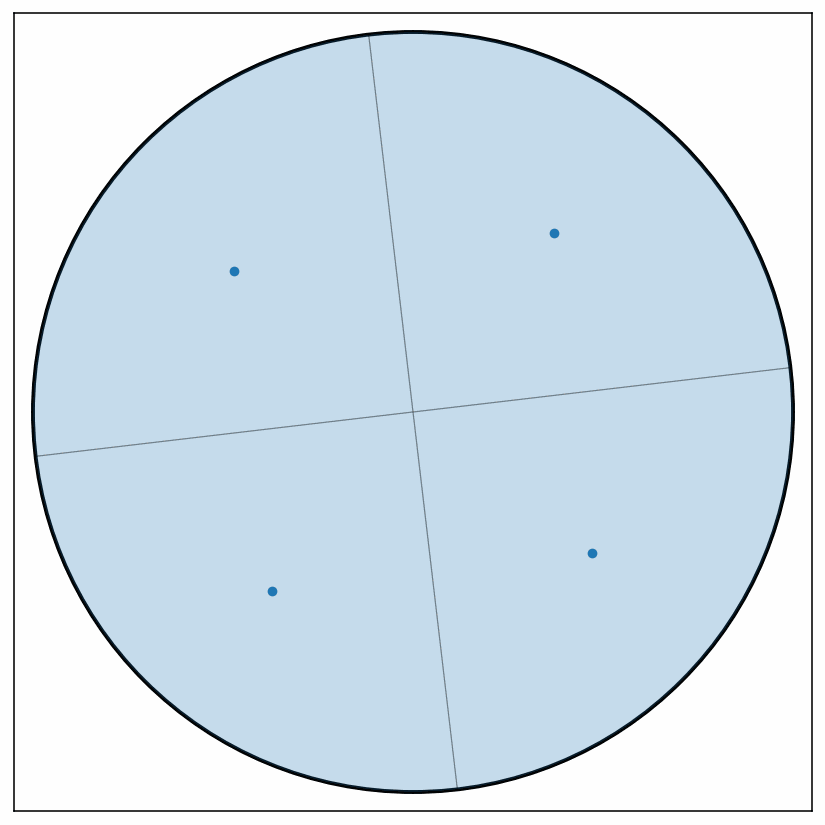}
        \caption{4 agents}
        \label{fig:a}
    \end{subfigure}
    \hfill
    \begin{subfigure}{0.48\columnwidth}
        \centering
        \includegraphics[width=\linewidth]{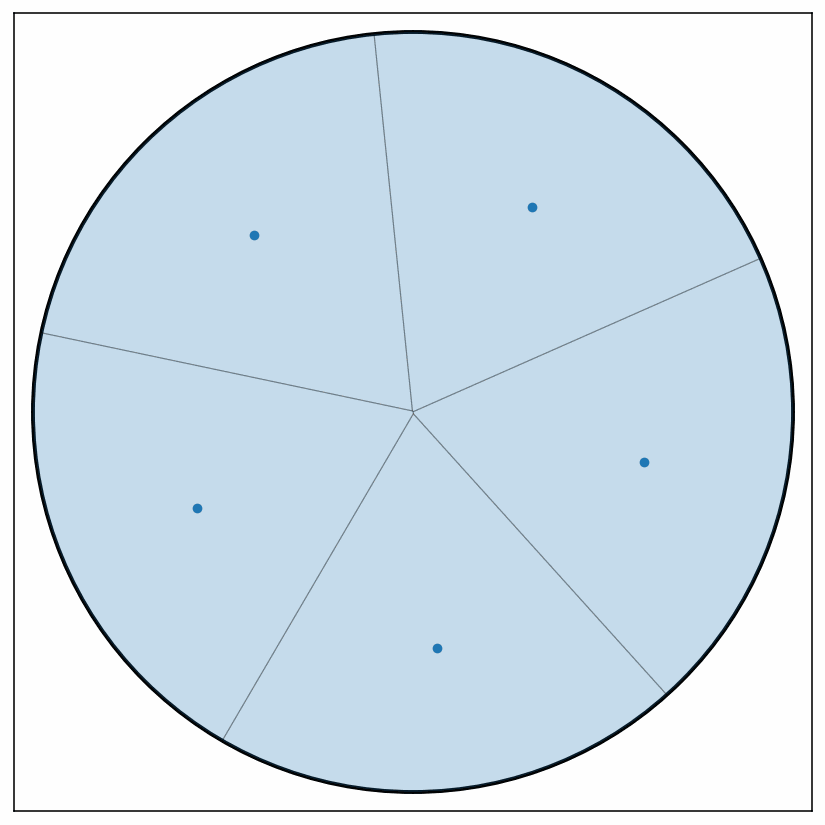}
        \caption{5 Agents}
        \label{fig:b}
    \end{subfigure}
    
    \vspace{0.5em}
    
    % Second row
    \begin{subfigure}{0.48\columnwidth}
        \centering
        \includegraphics[width=\linewidth]{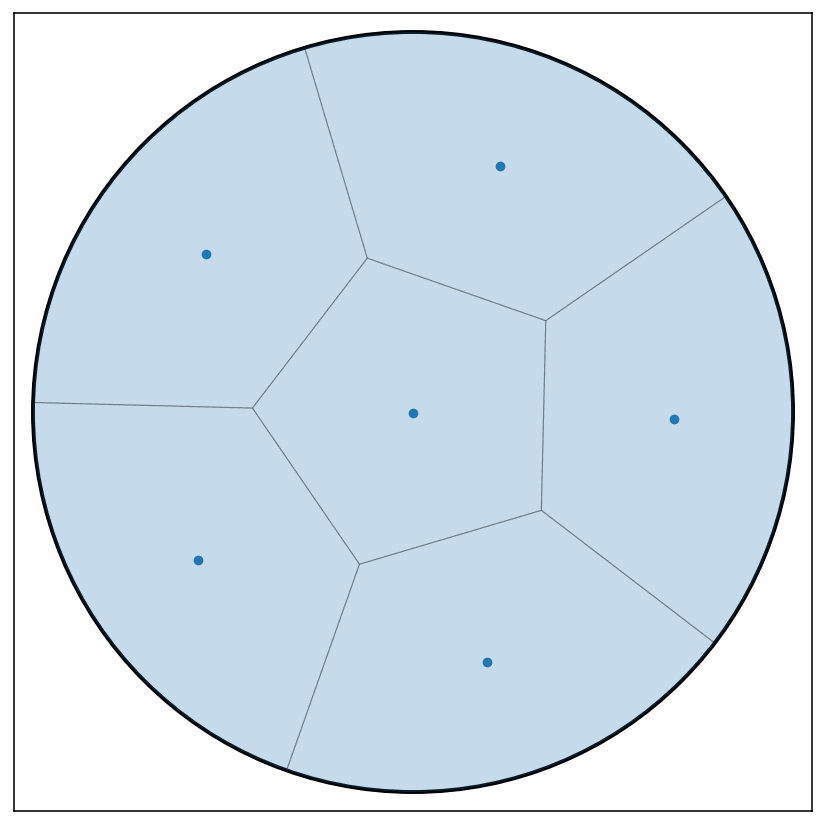}
        \caption{6 Agents}
        \label{fig:c}
    \end{subfigure}
    \hfill
    \begin{subfigure}{0.48\columnwidth}
        \centering
        \includegraphics[width=\linewidth]{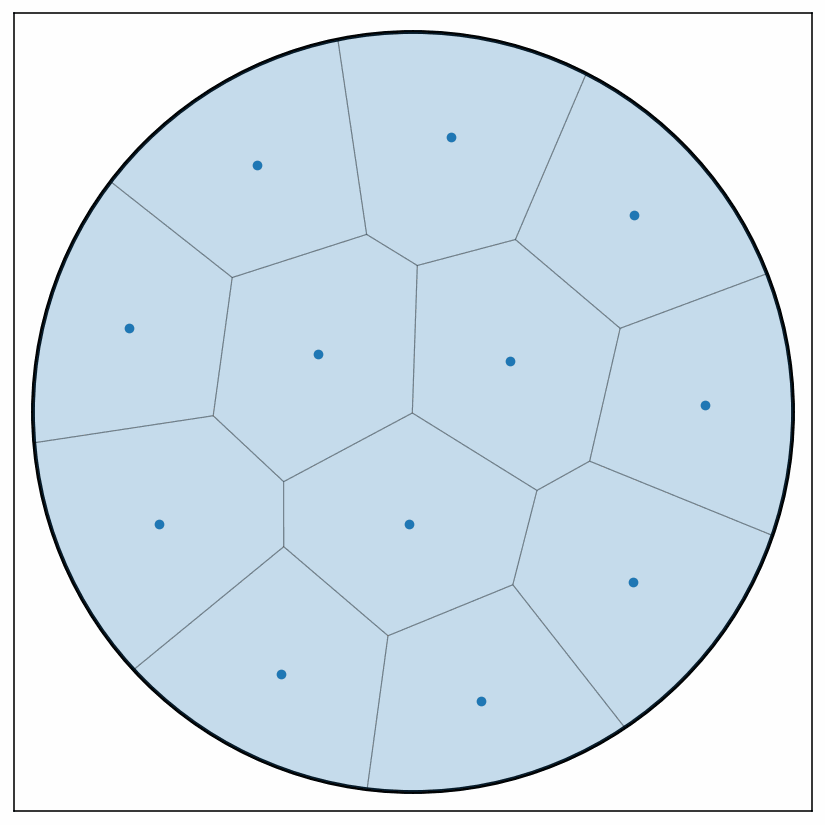}
        \caption{12 Agents}
        \label{fig:d}
    \end{subfigure}
    
    \caption{Optimized agent separation with Voronoi partitions after 20 iterations of  Lloyd's relaxation, constrained to a circular boundary for 4, 5, 6, and 12 agents.}
    \label{fig:voronoi}
\end{figure}

\subsection{Baseline: Voronoi Partitioning with Lloyd's Relaxation}

To evaluate the effectiveness of our QMIX approach, we compare it against a baseline controller using Voronoi partitioning with Lloyd's relaxation. Voronoi partitioning divides an area into subregions given an initial configuration of starting points, where the boundaries of each subregion are equidistant to the nearest point.  Applying Lloyd's relaxation iteratively improves the geometry of the partition by moving the starting point toward the centroid of its assigned Voronoi cell, converging to a more uniform spatial distribution that minimizes coverage gaps and redundancy, typically resulting in a hexagonal tessellation pattern. Figure \ref{fig:voronoi} shows four examples of optimized Voronoi partitions constrained to a circular region for different numbers of agents after applying Lloyd's relaxation. 

The baseline controller, therefore, acts as an adaptive waypoint assigner to the centroids of iteratively optimized Voronoi partitions for optimal separation. We apply a simple greedy control method to the individual HAB agents, where the HAB adjusts its altitude to best drift in the direction of the waypoint.  Figure \ref{fig:baseline_time} shows how the baseline controller adapts the Voronoi partitions and re-assigns waypoints as the HABs drift.  It is important to note that this baseline controller does not take into account reachability constraints when assigning waypoints and is only distance-based.  A Voronoi partition-based baseline is particularly well-suited for distributed area coverage tasks as it provides theoretically optimal spatial partitioning for static scenarios and has been widely used in multi-agent coverage literature, including being proposed as a method for realizing global coverage with a HAB constellation \cite{du2022dynamic}. 
%While this deterministic approach cannot adapt to complex environmental dynamics or learn from experience like MARL methods, it provides a strong benchmark representing optimal geometric coverage under idealized conditions.

\begin{table*}[!t]
\centering
\caption{Local Agent Observation Space Components}
\label{tab:local_observations}
\renewcommand{\arraystretch}{1.4}
\setlength{\tabcolsep}{8pt}
\begin{tabular}{p{3.5cm}p{2.5cm}p{3cm}p{5.5cm}}
\hline
\textbf{Feature} & \textbf{Range} & \textbf{Normalized Range} & \textbf{Notes} \\
\hline
\hline
\multicolumn{4}{c}{\textbf{Environmental Context}} \\
\hline
Wind profile ($\Phi_i$) & Variable & [0, 1]$^{37 \times 3}$ & Multi-level wind data: altitude, bearing, velocity \\
Shared goal ($goal_{shared}$) & $\pm$150 km & [0, 1]$^2$ & Normalized coordinates of team coverage center \\
\hline
\multicolumn{4}{c}{\textbf{Team Coordination}} \\
\hline
Other agents ($others_i$) & Variable & [0, 1]$^{5 \times (N-1)}$ & Positions, distances, altitudes, goal distances of teammates \\
\hline
\multicolumn{4}{c}{\textbf{Individual State Information}} \\
\hline
Altitude ($alt_i$) & 15--25 km & [0, 1] & Normalized within operational bounds \\
Position ($pos_i$) & $\pm$150 km & [0, 1]$^2$ & Agent coordinates normalized by max operational distance \\
Coverage membership ($w_{goal,i}$) & Boolean & \{0, 1\} & Binary flag indicating coverage area membership \\
Distance to goal ($d_{goal,i}$) & 0--300 km & [0, 2] & Normalized by coverage radius; can exceed 1.0 when outside \\
Relative bearing ($\theta_{goal,i}$) & 0--2$\pi$ rad & [0, 1] & Normalized relative bearing to coverage center \\
\hline
\end{tabular}
\begin{flushleft}
\textbf{Total observation dimension}: $6 + 3 \times N_{levels} + 5 \times (N_{agents} - 1) = 127$ dimensions for $N_{levels}=37$, $N_{agents}=3$
\end{flushleft}
\end{table*}

\subsection{Multi-Agent Observation and State Space Design}

The design of observation and state spaces in cooperative MARL for distributed area coverage must address the fundamental challenge of the credit assignment problem that arises when transitioning from individual to shared rewards. When agents receive shared team rewards, individual agents may receive positive reinforcement even when their individual behavior is suboptimal, even if only one other team member is performing well or vice versa. This creates a critical need for comprehensive local and global observation spaces that enable agents to correctly interpret shared global rewards in the context of their current individual states.

\subsubsection{Local Agent Observations}

Our Centralized Training with Decentralized Execution (CTDE) framework employs carefully designed observation spaces that provide each agent with sufficient information to understand their current individual state and contribution to team performance, despite receiving a global shared reward. Each agent $i$ receives a local observation $o_i$ formally defined as:

\begin{equation}
o_i = \begin{cases}
\text{Environmental Context:} & \{\Phi_i, goal_{shared}\} \\
\text{Team Coordination:} & \{others_i\}\\
\text{Individual State:} & \{alt_i, pos_i, d_{goal,i}, \\
& \quad w_{goal,i}, \theta_{goal,i}\} \\
\end{cases}
\end{equation}

The observation space shown in Table~\ref{tab:local_observations} is structured to address specific coordination challenges across three categories: individual state information provides each agent with its current position and coverage status; environmental context includes multi-level wind profiles and shared team objectives; team coordination information enables agents to understand teammate states and positions. The inclusion of explicit coverage status flags ($w_{goal,i}$) and comprehensive teammate information enables individual agents to understand if they are contributing to the overall team performance, despite receiving identical global rewards. This rich local information is essential for stable learning under shared reward structures to guide agents in suboptimal states towards the target coverage area.

\subsubsection{Global State Space}

The QMIX mixing network receives a comprehensive global state $s$ for optimal coordination decisions during centralized training:

\begin{equation}
s = \begin{cases}
\text{Per-Agent States:} & \{s_1, s_2, ..., s_{N_{agents}}\} \\
\text{Shared Information:} & \{goal_{shared}\} \\
\text{Team Metrics:} & \{metrics_{team}\}
\end{cases}
\end{equation}

\begin{table*}[!t]
\centering
\caption{Global State Space Components}
\label{tab:global_state}
\renewcommand{\arraystretch}{1.4}
\setlength{\tabcolsep}{8pt}
\begin{tabular}{p{3.5cm}p{2.5cm}p{3cm}p{5.5cm}}
\hline
\textbf{Feature} & \textbf{Range} & \textbf{Normalized Range} & \textbf{Notes} \\
\hline
\hline
\multicolumn{4}{c}{\textbf{Per-Agent State ($\times N_{agents}$)}} \\
\hline
Agent altitude & 15--25 km & [0, 1] & Individual agent altitude \\
Agent position & $\pm$150 km & [0, 1]$^2$ & Individual agent coordinates \\
Distance to goal & 0--300 km & [0, 2] & Individual distance to shared goal \\
Coverage status & Boolean & \{0, 1\} & Individual coverage area status \\
Relative bearing & 0--2$\pi$ rad & [0, 1] & Individual bearing to goal \\
Wind profile & Variable & [0, 1]$^{37 \times 3}$ & Individual agent's wind column \\
\hline
\multicolumn{4}{c}{\textbf{Shared Information}} \\
\hline
Team goal & $\pm$150 km & [0, 1]$^2$ & Shared coverage center coordinates \\
\hline
\multicolumn{4}{c}{\textbf{Team Metrics}} \\
\hline
Coverage ratio & 0--100\% & [0, 1] & Fraction of agents within coverage area \\
Separation score & 0--1 & [0, 1] & Normalized measure of agent dispersion \\
\hline
\end{tabular}
\begin{flushleft}
\textbf{Total global state dimension}: $N_{agents} \times (6 + 3 \times N_{levels}) + 4 = 355$ dimensions for $N_{agents}=3$, $N_{levels}=37$
\end{flushleft}
\end{table*}

The global state shown in Table~\ref{tab:global_state} includes complete information for each agent, shared team objectives, and team-level coordination metrics that inform the mixing network about overall coordination effectiveness. This comprehensive state representation provides the mixer with complete observability while maintaining the CTDE paradigm requirements of QMIX.

\subsection{Action Space and Dynamics}

Each agent maintains the same discrete action space as the single-agent system, with three altitude control actions based on SHAB-V flight characteristics:

\begin{equation}
a_i \in \{ASCEND, MAINTAIN, DESCEND\}
\end{equation}

These actions map to altitude change distributions:
\begin{itemize}
    \item ASCEND: $\mathcal{N}(1.80, 0.14^2)$ m/s
    \item MAINTAIN: $\mathcal{N}(0.00, 1.25^2)$ m/s  
    \item DESCEND: $\mathcal{N}(-2.80, 0.30^2)$ m/s
\end{itemize}

\subsection{Cooperative Reward Function Design}

%The transition from individual to shared rewards creates unique design requirements that distinguish multi-agent HAB coordination from single-agent station-keeping tasks. 
In our previous single-agent work, agents received individual rewards based solely on their relative distance and bearing to the station center target region using a piecewise reward function. This approach works well for independent agents with an individual station-keeping objective, but fails fundamentally for cooperative area coverage tasks, where:

\textbf{Spatial Distribution Matters-} Individual success in assigned positions does not guarantee optimal global coverage if agents cluster together.

\textbf{Dynamic Reallocation is Required-} Agents must coordinate among each other to adaptively reposition based on atmospheric conditions and teammate states.

\textbf{Team Performance is Non-Decomposable-} Overall area coverage success cannot be expressed as a simple sum of individual performance metrics.

\subsubsection{Multi-Agent Reward Structure}

Our cooperative reward function addresses these limitations by implementing shared team-level objectives:

\begin{equation}
R_{team} = 10.0 \times R_{coverage} + 3.0 \times R_{dispersion}
\end{equation}

\textbf{Coverage Reward (Primary):}
The primary reward component incentivizes agents to remain within the target coverage area:

\begin{equation}
R_{coverage} = \frac{N_{within}}{N_{agents}}
\label{coverage_ratio}
\end{equation}

where $N_{within}$ is the number of agents currently within the coverage radius.

\textbf{Dispersion Reward (Secondary):}
The secondary reward component encourages optimal spatial distribution, but only applies when multiple agents are within the coverage area:

\begin{equation}
R_{dispersion} = \begin{cases}
\min\left(\frac{\bar{d}_{separation}}{d_{target}}, 1.0\right) & \text{if } N_{within} \geq 2 \\
0 & \text{otherwise}
\end{cases}
\label{dispersion_reward}
\end{equation}

\begin{equation}
d_{target} = R_{coverage}/\sqrt{N_{agents}}
\label{separation_ratio}
\end{equation}

where $d_{target}$ is the target separation distance and $\bar{d}_{separation}$ is the average pairwise distance between agents inside the coverage area.

\subsubsection{Design Rationale and Credit Assignment Solution}

The shared reward structure creates the fundamental challenge that agents may receive high global rewards even when their individual states are suboptimal. Our solution involves three complementary design elements:

\textbf{Hierarchical Reward Priority:} The 10:3 ratio between coverage and dispersion rewards ensures that staying within the target area takes absolute priority over optimal agent distribution.

\textbf{Conditional Dispersion Rewards:} Dispersion rewards only apply when $N_{within} \geq 2$, ensuring that spatial coordination incentives do not conflict with the primary station-keeping coverage objective.

\textbf{Rich Local Observation Integration:} The explicit coverage status flags in local observations enable individual networks to correctly interpret global rewards. An agent outside the coverage area can recognize its negative contribution despite receiving positive team rewards to guide itself towards the shared coverage area.

This integrated design ensures that the Individual-Global-Max (IGM) principle is satisfied while providing sufficient information density to enable effective coordination learning under shared reward structures.

\subsection{QMIX Architecture Configuration}

Our QMIX implementation consists of individual agent Q-networks and a mixing network:

\begin{itemize}
    \item \textbf{Agent Networks}: 4-layer fully connected networks (256 hidden units each) mapping observations to Q-values for 3 actions
    \item \textbf{Mixing Network}: Hypernetwork-based mixer with 64-dimensional embedding, taking individual Q-values and global state to produce $Q_{tot}$
    \item \textbf{Training Parameters}: Learning rate $1\times10^{-6}$, batch size 128, replay buffer size $10^6$, $\epsilon$-greedy exploration decaying from 1.0 to 0.05 over 2M steps
\end{itemize}

\begin{figure}[!t] % stays within one column
    \centering
    
    % First row
    \begin{subfigure}{\columnwidth}
        \centering
        \includegraphics[width=\linewidth]{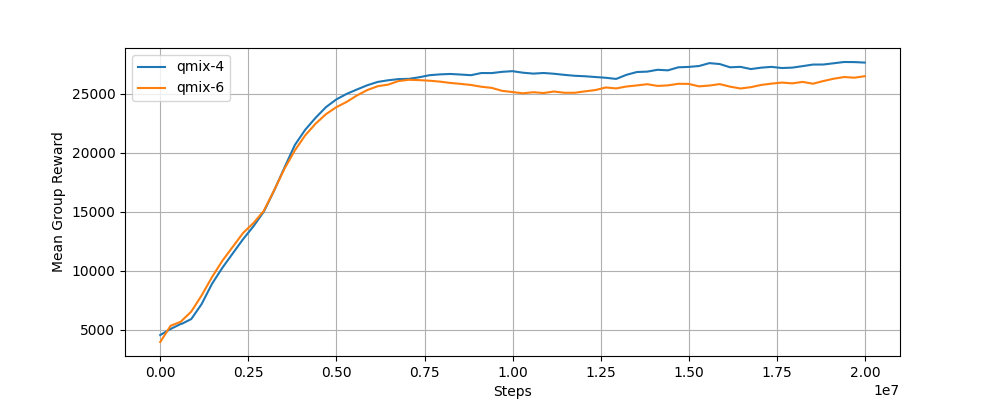}
        \caption{Mean Reward}
        \label{fig:a}
    \end{subfigure}
    
    \vspace{0.5em}
    
    % Second row
    \begin{subfigure}{\columnwidth}
        \centering
        \includegraphics[width=\linewidth]{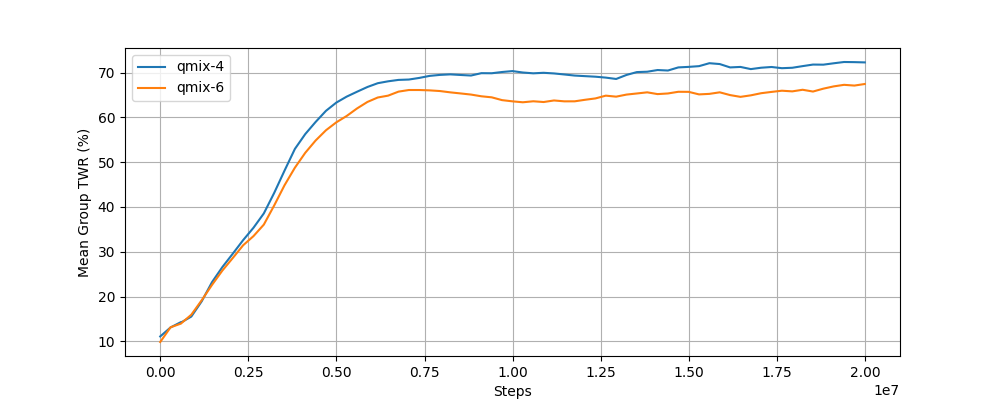}
        \caption{Mean Group TWR}
        \label{fig:b}
    \end{subfigure}

    \vspace{0.5em}
    
    % Second row
    \begin{subfigure}{\columnwidth}
        \centering
        \includegraphics[width=\linewidth]{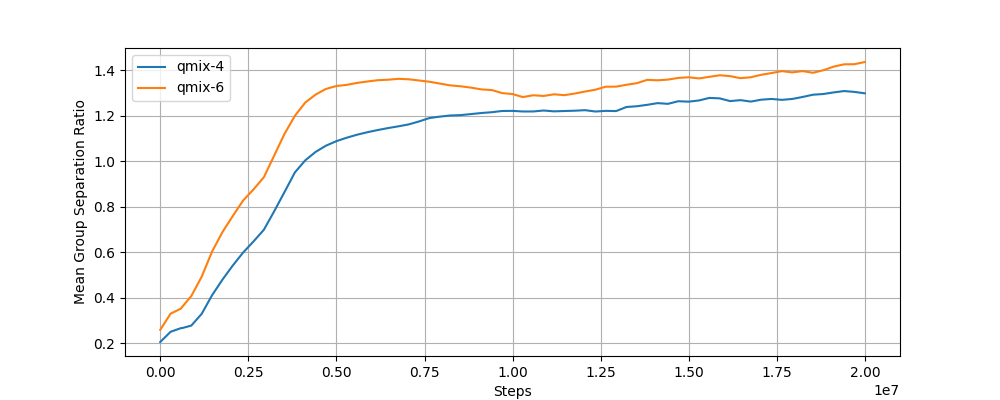}
        \caption{Mean Separation Ratio}
        \label{fig:c}
    \end{subfigure}
    \hfill
    
    \caption{QMIX Learning Curves for Reward, Separation Ratio, and Mean Group TWR over 7000 episodes (approximately 20 million timesteps) for 4 and 6 agents.}
    \label{fig:training}
\end{figure}

% \begin{figure*}[!t] % spans both columns
%     \centering
    
%     \begin{subfigure}{0.32\textwidth}
%         \centering
%         \includegraphics[width=\linewidth]{img/learning_reward.png}
%         \caption{Mean Reward}
%         \label{fig:a}
%     \end{subfigure}
%     \hfill
%     \begin{subfigure}{0.32\textwidth}
%         \centering
%         \includegraphics[width=\linewidth]{img/learning_twr.png}
%         \caption{Mean Group TWR}
%         \label{fig:b}
%     \end{subfigure}
%     \hfill
%     \begin{subfigure}{0.32\textwidth}
%         \centering
%         \includegraphics[width=\linewidth]{img/learning_separation.png}
%         \caption{Mean Separation Ratio}
%         \label{fig:c}
%     \end{subfigure}
    
%     \caption{QMIX Learning Curves for Reward, Separation Ratio, and Mean Group TWR over 7000 episodes (approximately 20 million timesteps) for 4 and 6 agents.}
%     \label{fig:training}
% \end{figure*}

%%%%%%%%%%%%%%%%%%%%%%%%%%%%%%%%%%%%%%
\section{Results and Analysis}
%%%%%%%%%%%%%%%%%%%%%%%%%%%%%%%%%%%%%%
We trained our QMIX approach using the same wind forecast setup as our previous single-agent reinforcement learning setup: ERA5 reanalysis data for observations and synthetic wind fields derived from radiosonde data for actual movement \cite{schuler2025seasonal}. All QMIX trainings occurred over 7,000 episodes (approximately 20 million timesteps) in the Southwestern United States region in the month of July, with randomized HAB starting positions within the circular coverage area and a wide variety of HAB-navigable wind conditions. These training runs revealed consistent convergence patterns across different team sizes: all tested training configurations (4, 6, and 11 agents) demonstrate stable learning with convergence occurring around 5 million timesteps, reaching final mean rewards of approximately 25,000-27,000 and mean group TWR values near 65-70\%. Figure~\ref{fig:training} shows representative QMIX training curves for the 4-agent and 6-agent configurations.

We compare the performance of QMIX against our baseline controller to assess the effectiveness of cooperative MARL for distributed area coverage tasks by evaluating two primary metrics:
%designed to capture the essential requirements of distributed area coverage:

\textbf{Time Within Region (TWR):} TWR measures the percentage of episode time that agents spend within the overall target coverage area. Higher TWR values indicate better coverage maintenance, with team-level TWR representing the average across all agents.

\textbf{Separation Score:} This metric quantifies spatial distribution quality by measuring the average pairwise distance between agents within the coverage area, normalized by the target separation distance $d_{target} = R_{coverage}/\sqrt{N_{agents}}$. Values near 1.0 indicate optimal spacing based on our optimal agent separation heuristic, while values significantly below 1.0 suggest clustering.

%\textbf{Supplementary Metrics:} We also track coverage ratio (fraction of agents within the target area at each timestep), percent area covered (spatial extent of the target region actually covered by agent sensing radii), and spatial distribution quality measures. Coverage entropy quantifies the diversity of visit patterns across the target area, with higher values indicating a more uniform coverage distribution. We compute spatial spread metrics and Moran's I spatial autocorrelation to assess geometric distribution and clustering patterns of covered areas. Additionally, we track individual agent coverage time to assess per-agent contribution to team objectives and computational efficiency metrics, including training convergence characteristics.

\begin{figure*}[!t]
    \centering

    % Row 1
    \begin{subfigure}{\textwidth}
        \centering
        \includegraphics[width=0.23\textwidth]{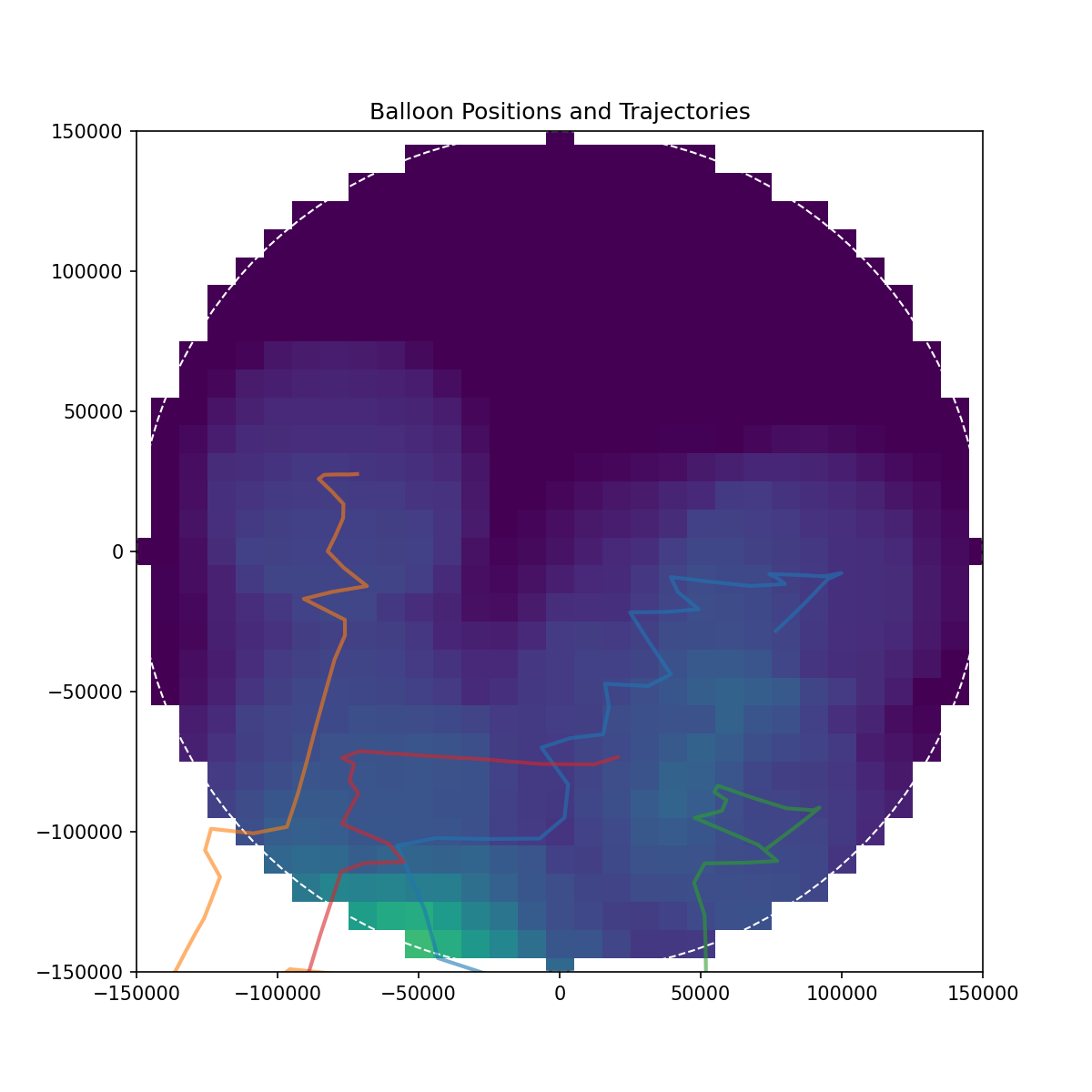}
        \includegraphics[width=0.23\textwidth]{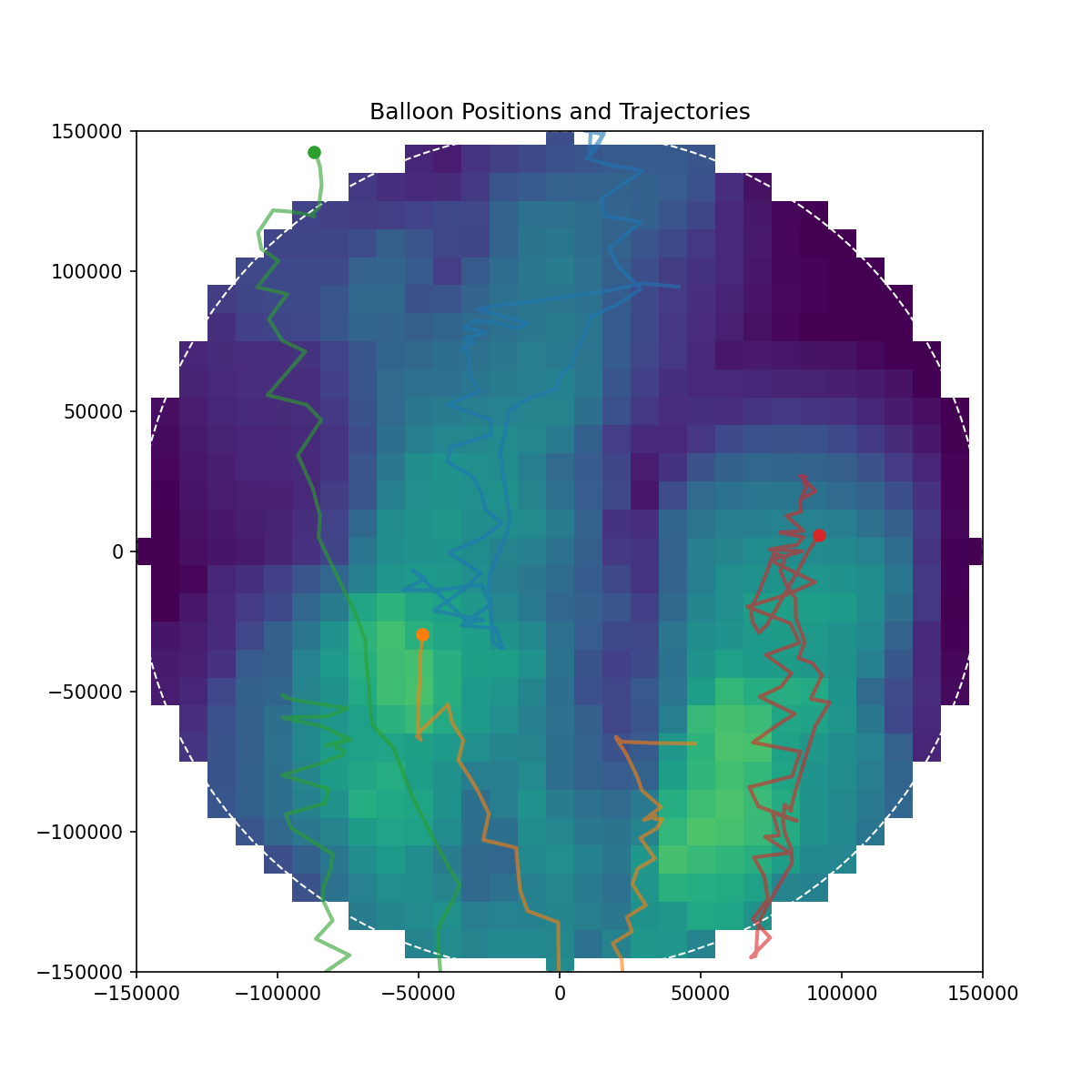}
        \includegraphics[width=0.23\textwidth]{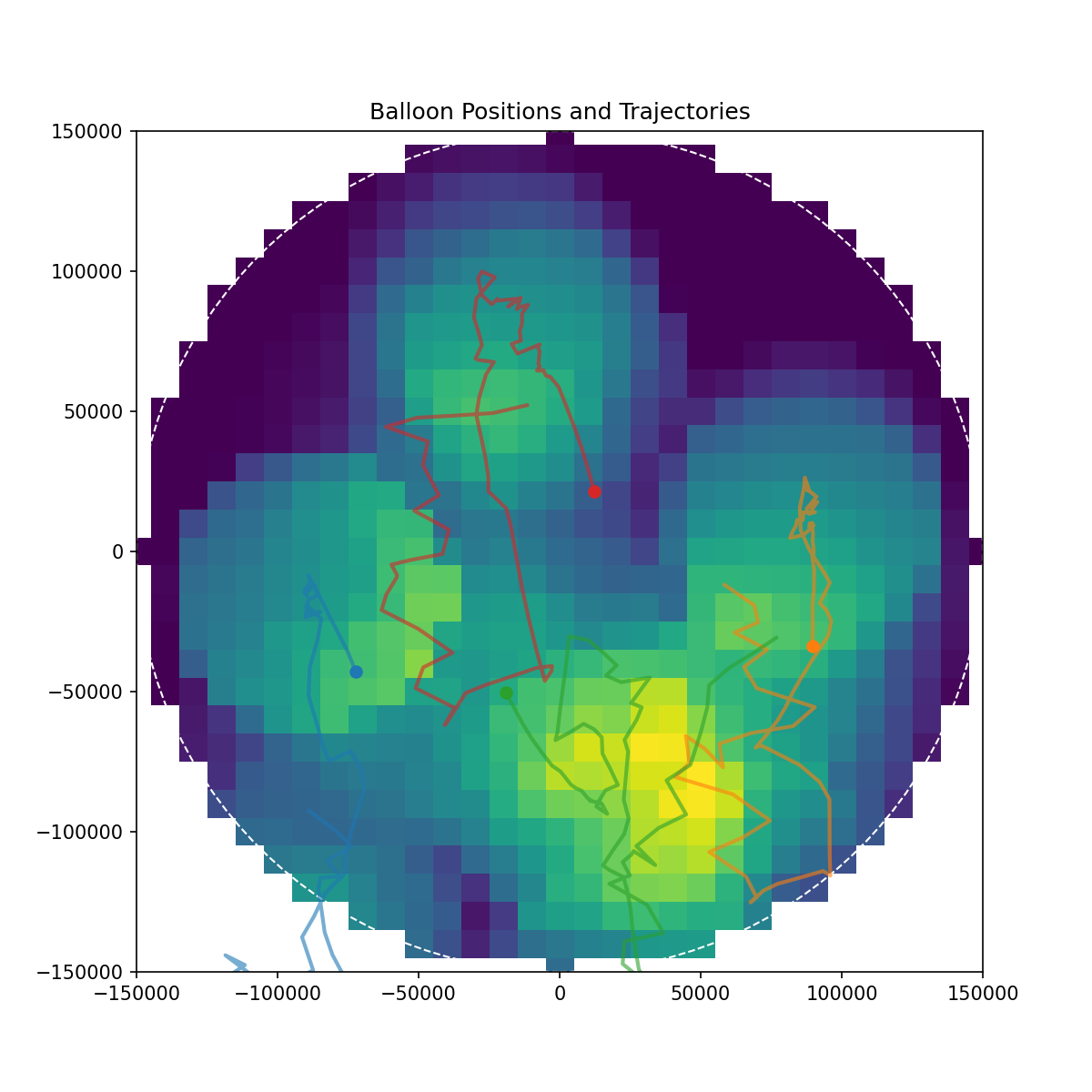}
        \includegraphics[width=0.23\textwidth]{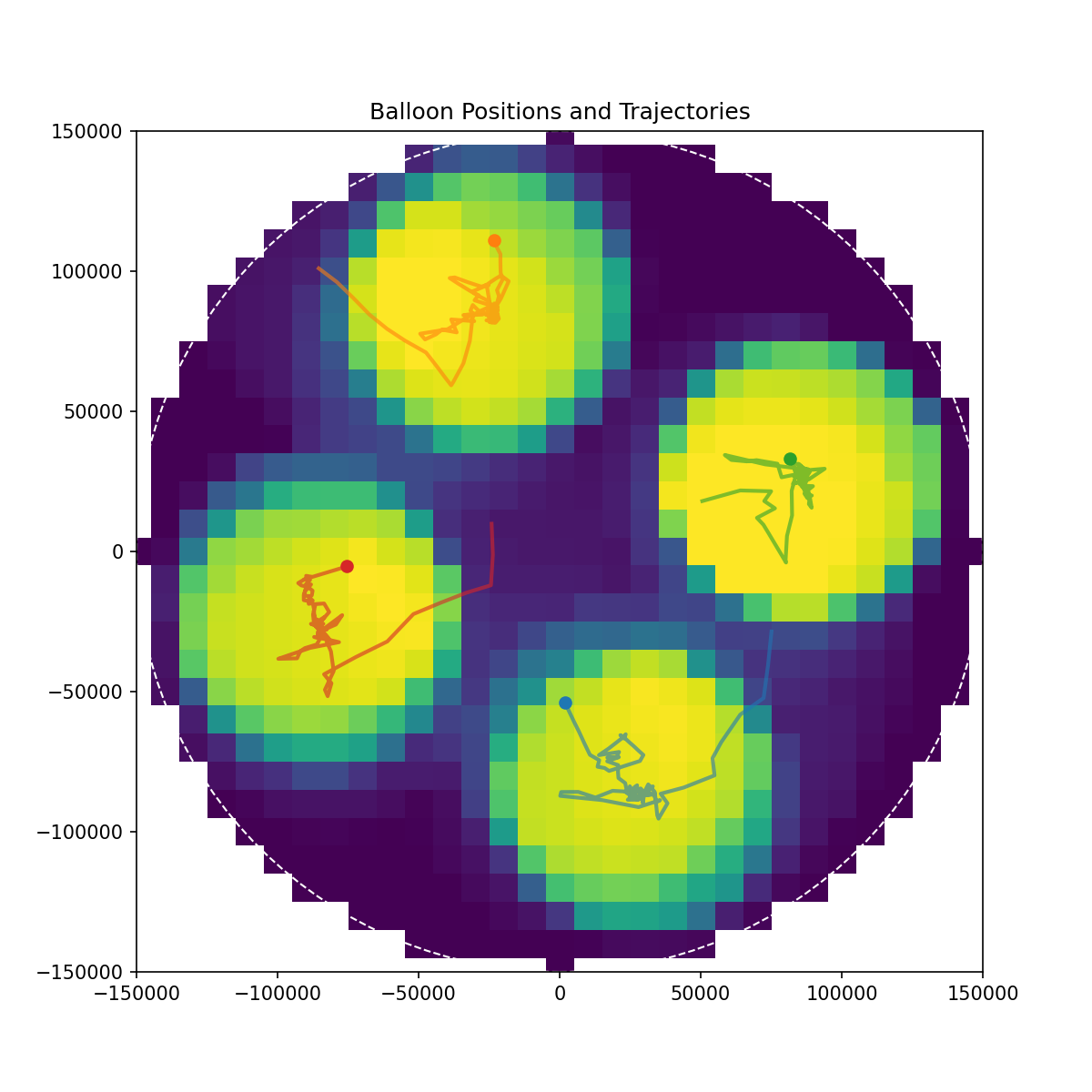}

        \caption{4-Agent Voronoi Baseline Controller}
        \label{fig:row1}
    \end{subfigure}

    \vspace{-0.1em}

    % Row 2
    \begin{subfigure}{\textwidth}
        \centering
        \includegraphics[width=0.23\textwidth]{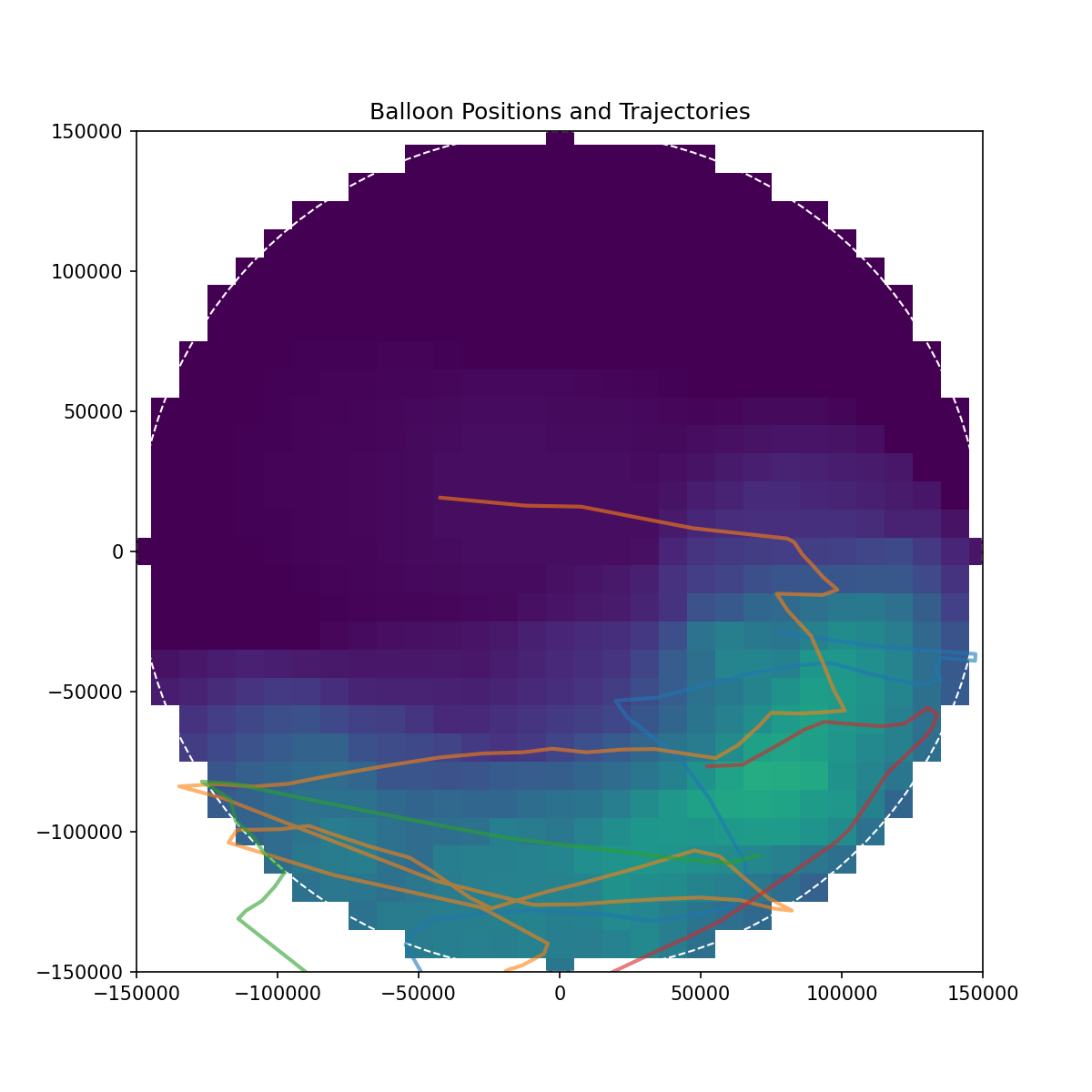}
        \includegraphics[width=0.23\textwidth]{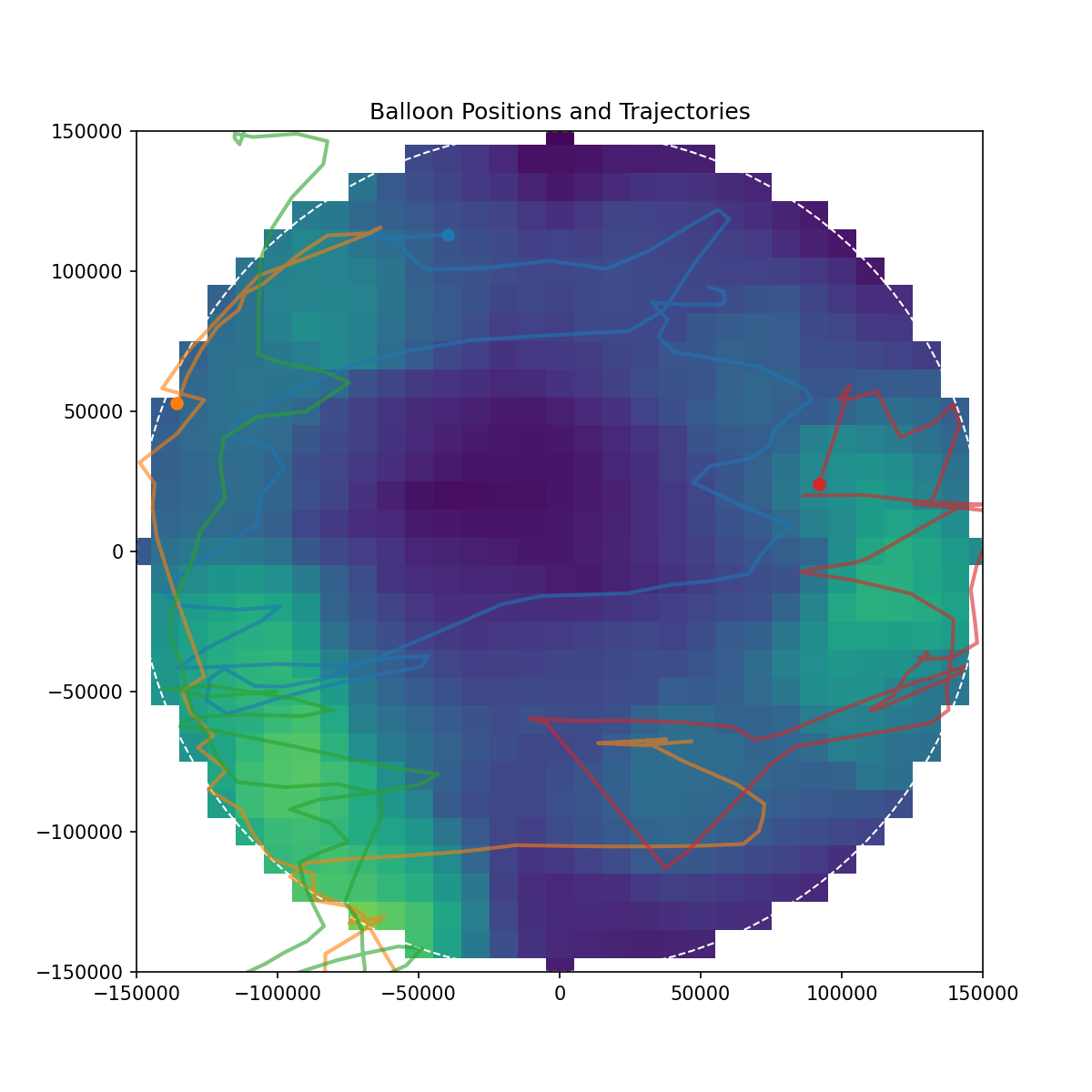}
        \includegraphics[width=0.23\textwidth]{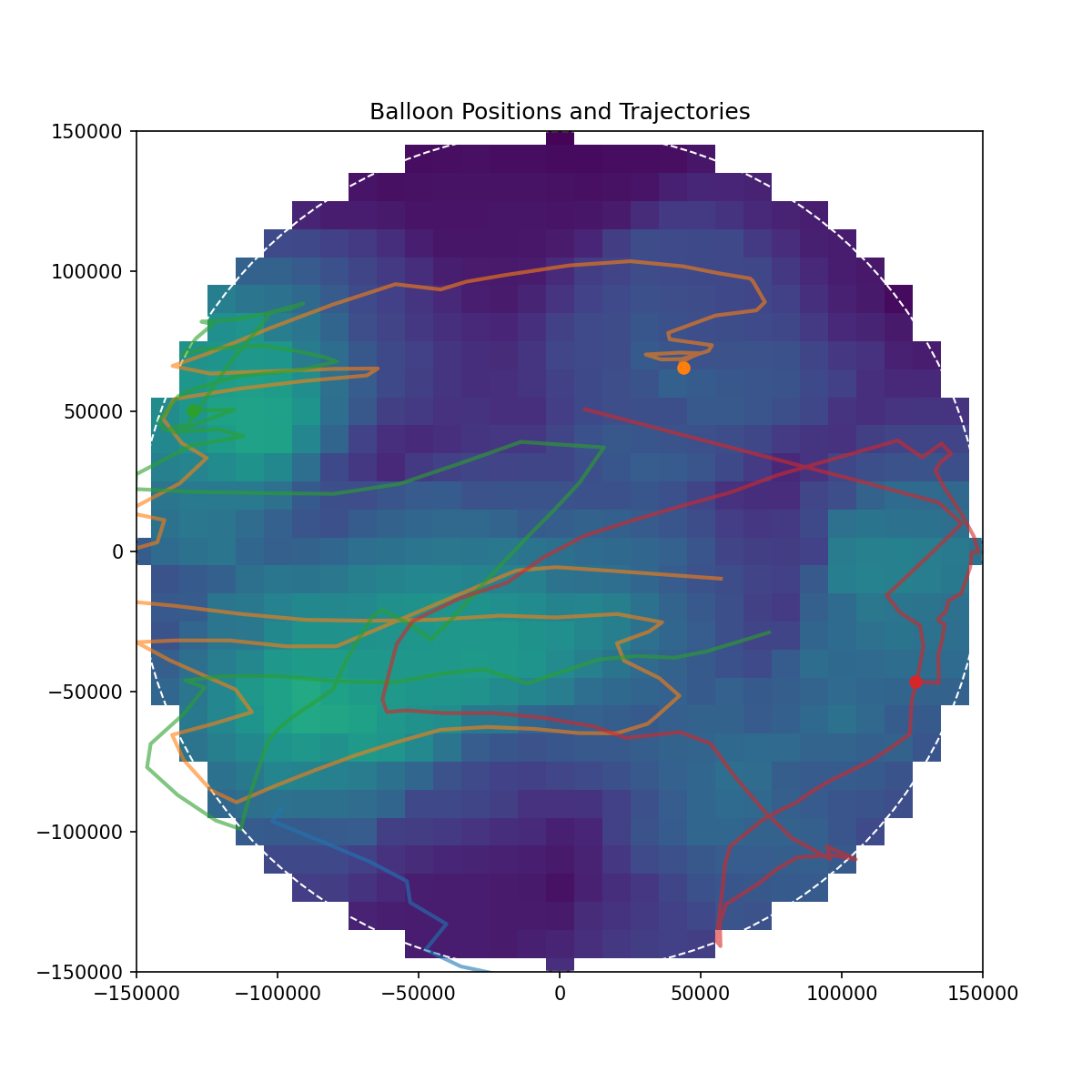}
        \includegraphics[width=0.23\textwidth]{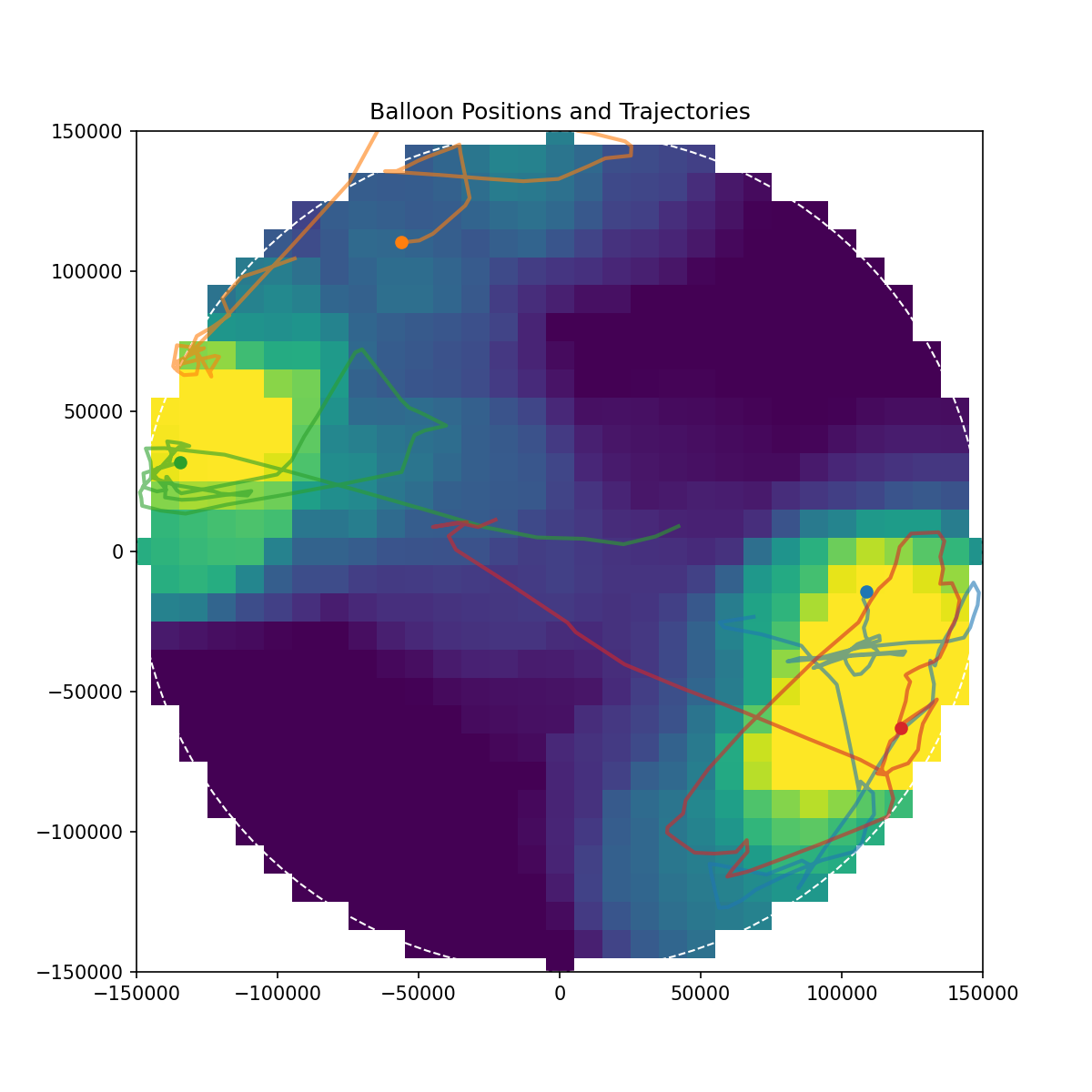}

        \caption{4-Agent QMIX Controller}
        \label{fig:row2}
    \end{subfigure}

    \vspace{-0.1em}

    % Row 3
    \begin{subfigure}{\textwidth}
        \centering
        \includegraphics[width=0.23\textwidth]{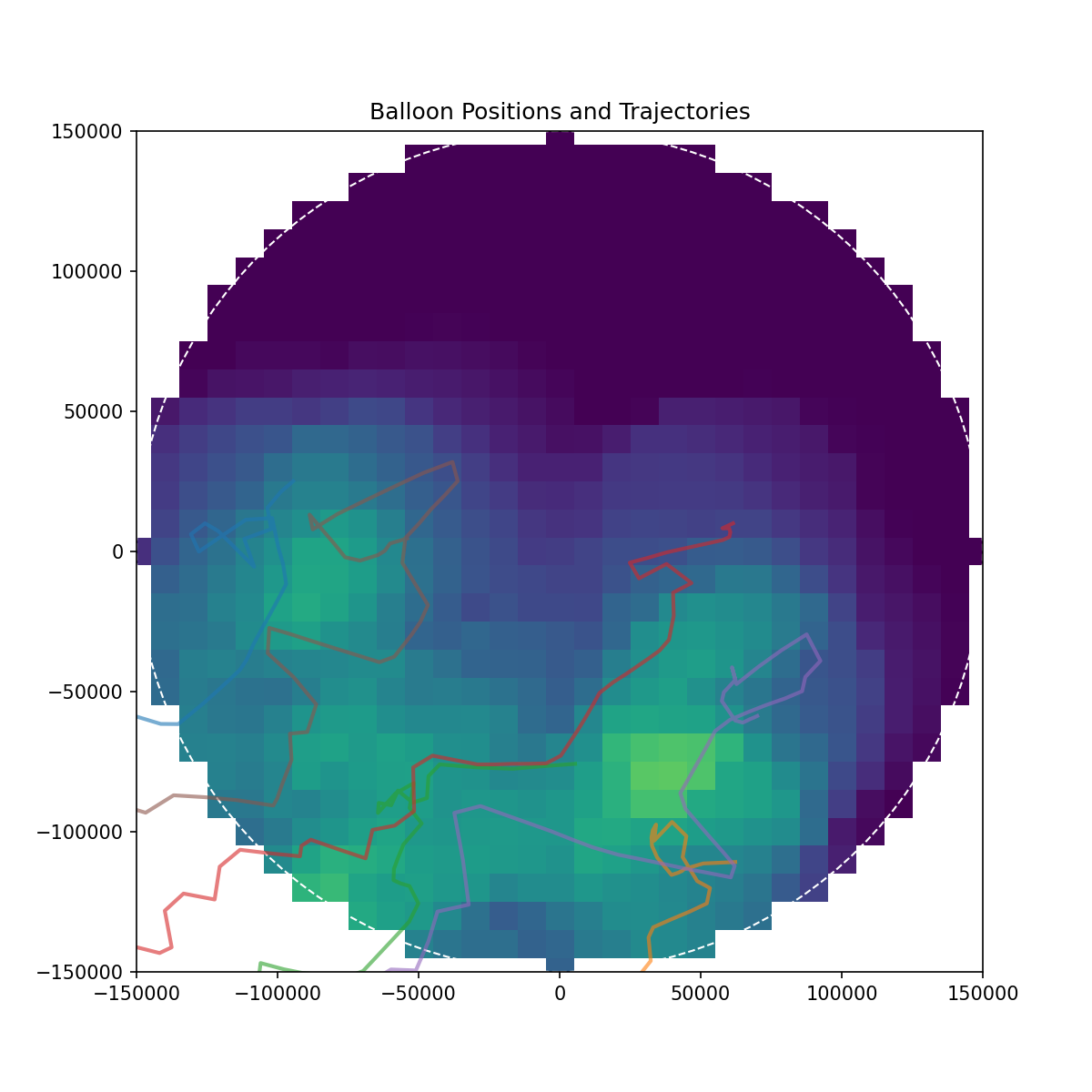}
        \includegraphics[width=0.23\textwidth]{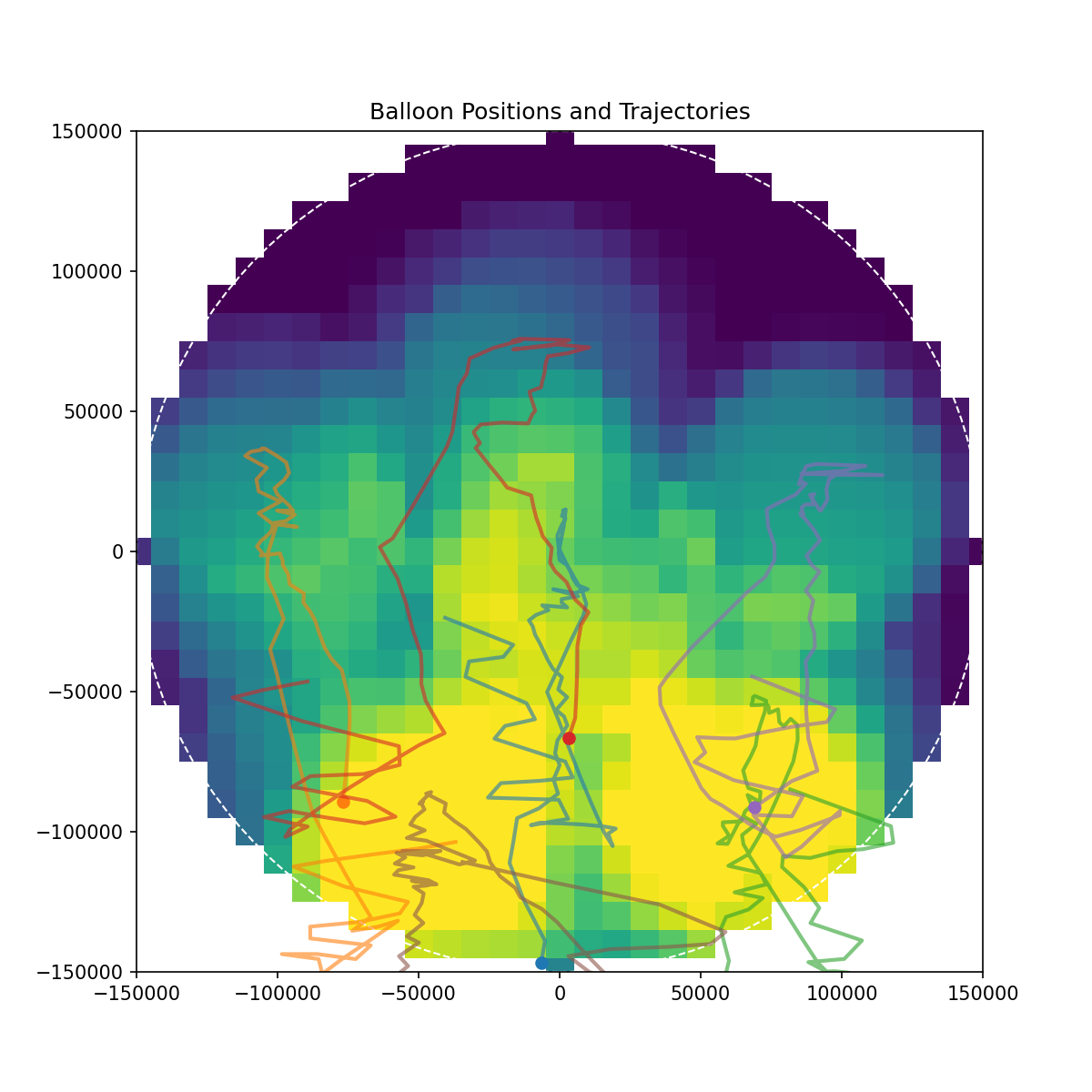}
        \includegraphics[width=0.23\textwidth]{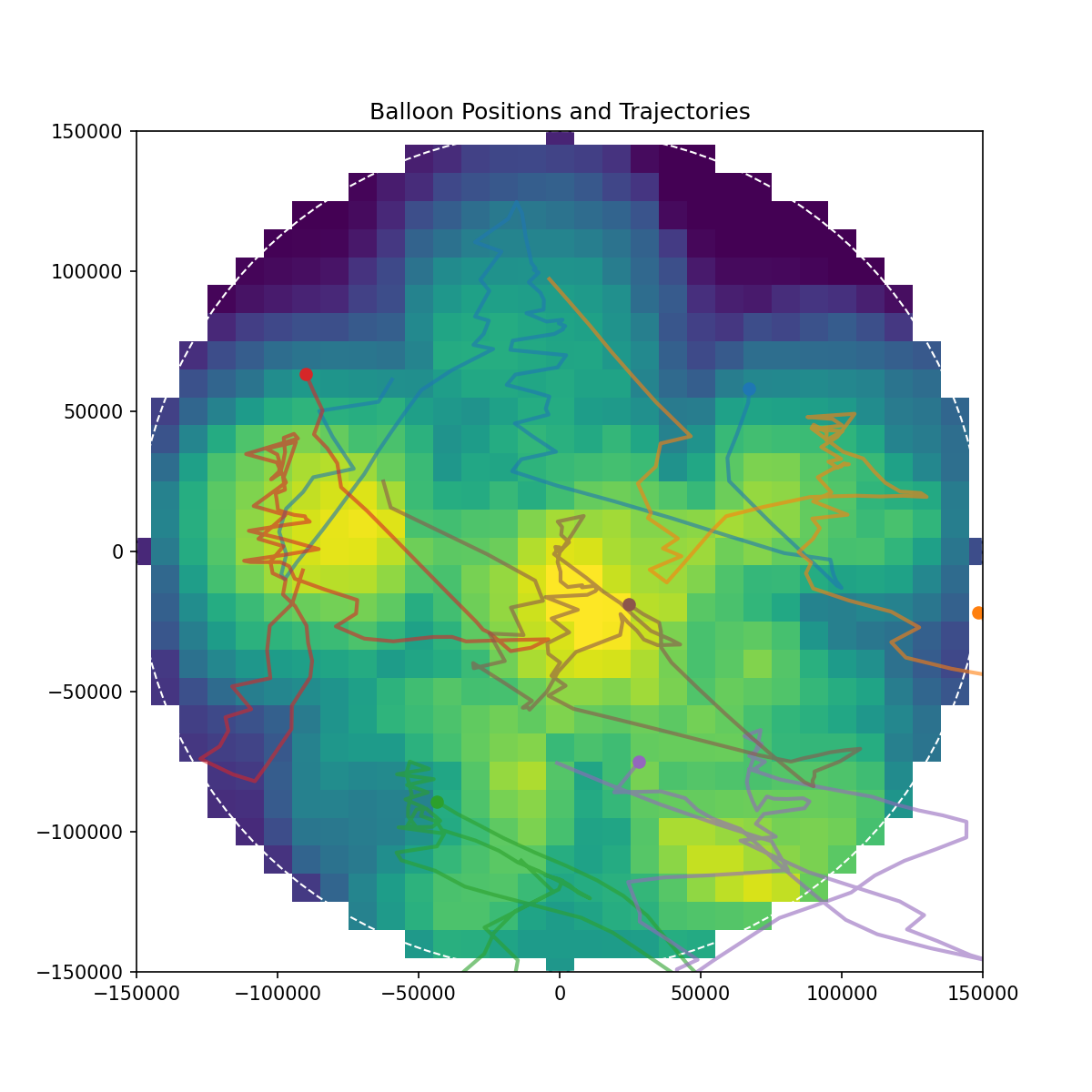}
        \includegraphics[width=0.23\textwidth]{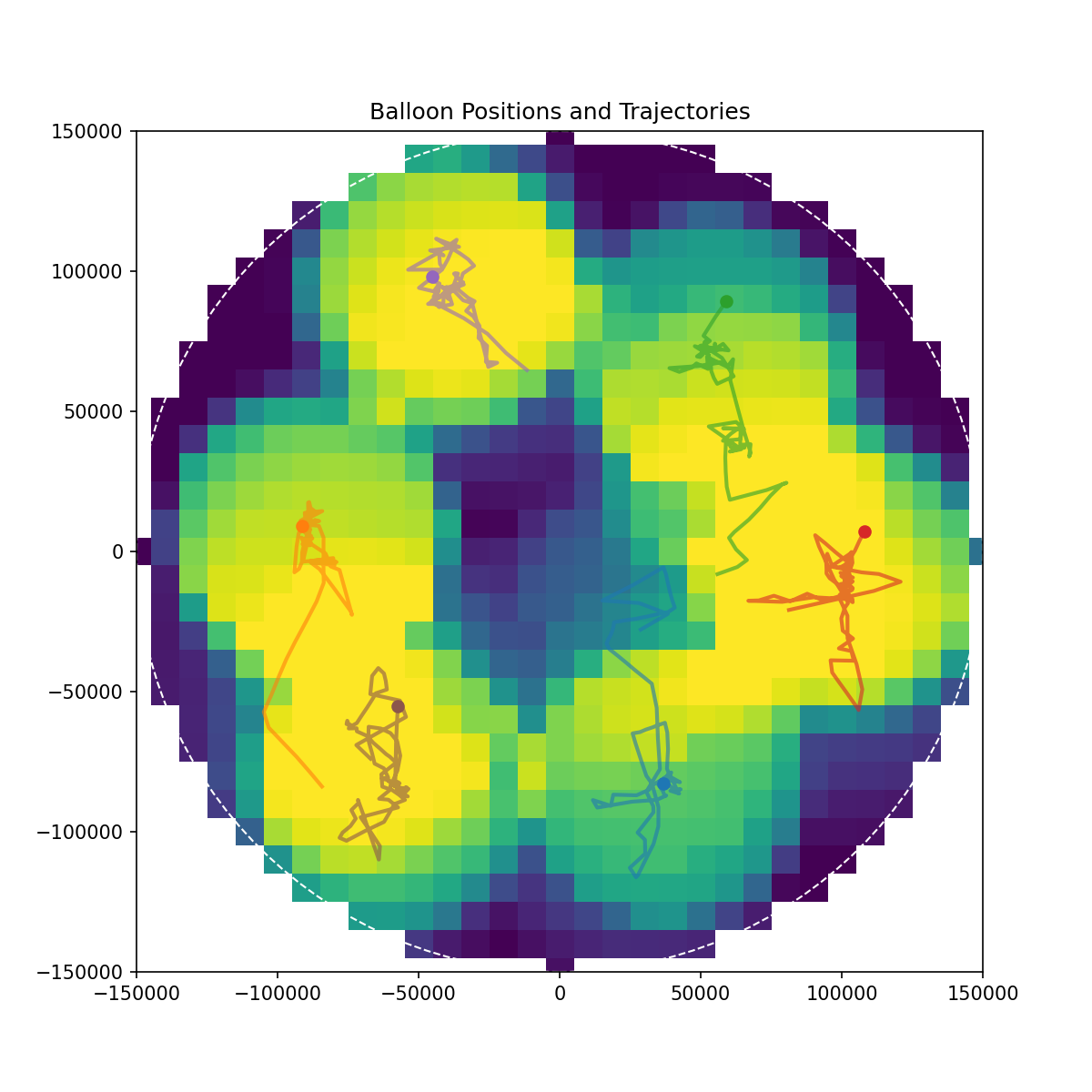}
        \caption{6-Agent Voronoi Baseline Controller}
        \label{fig:row3}
    \end{subfigure}

    \vspace{-0.1em}

    % Row 4
    \begin{subfigure}{\textwidth}
        \centering
        \includegraphics[width=0.23\textwidth]{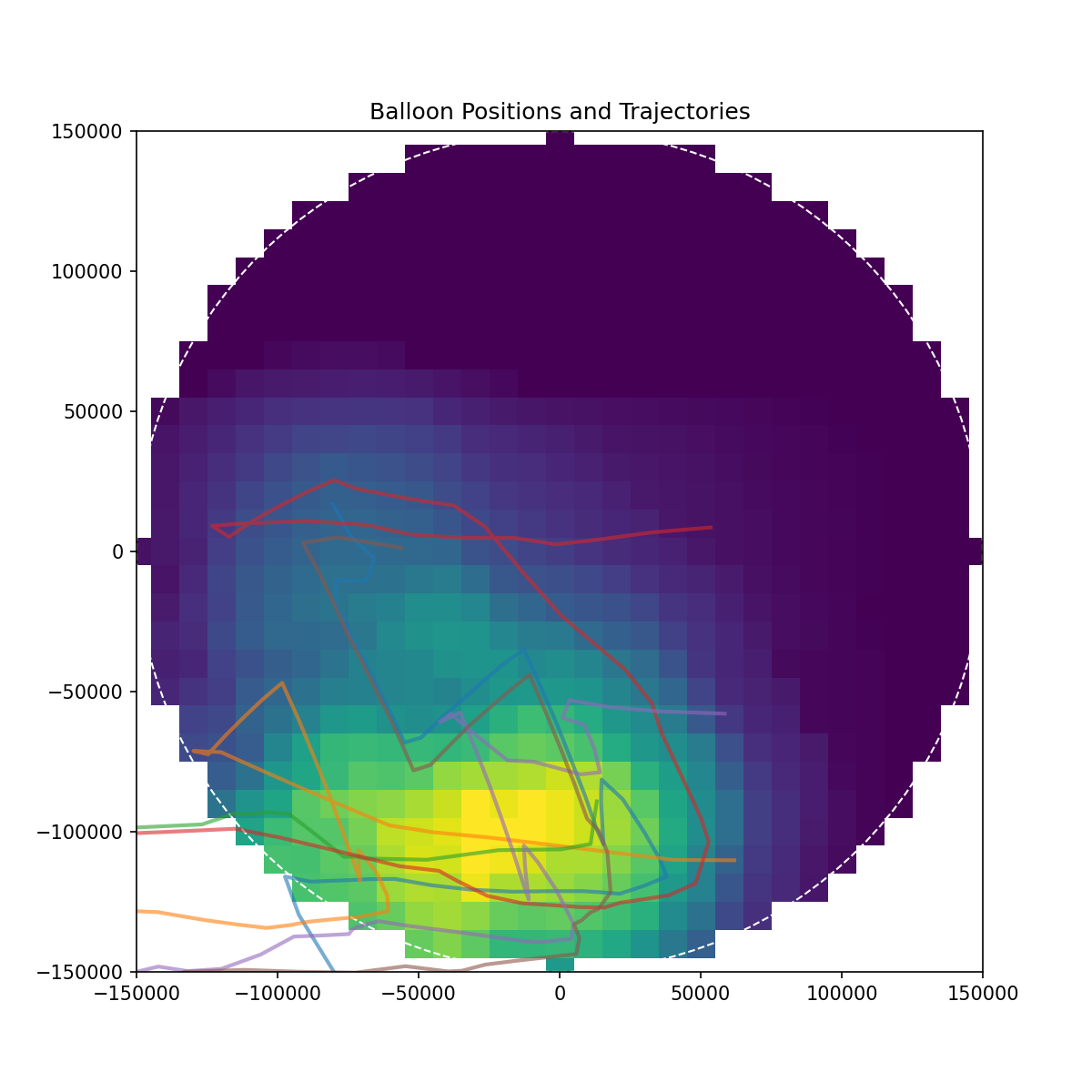}
        \includegraphics[width=0.23\textwidth]{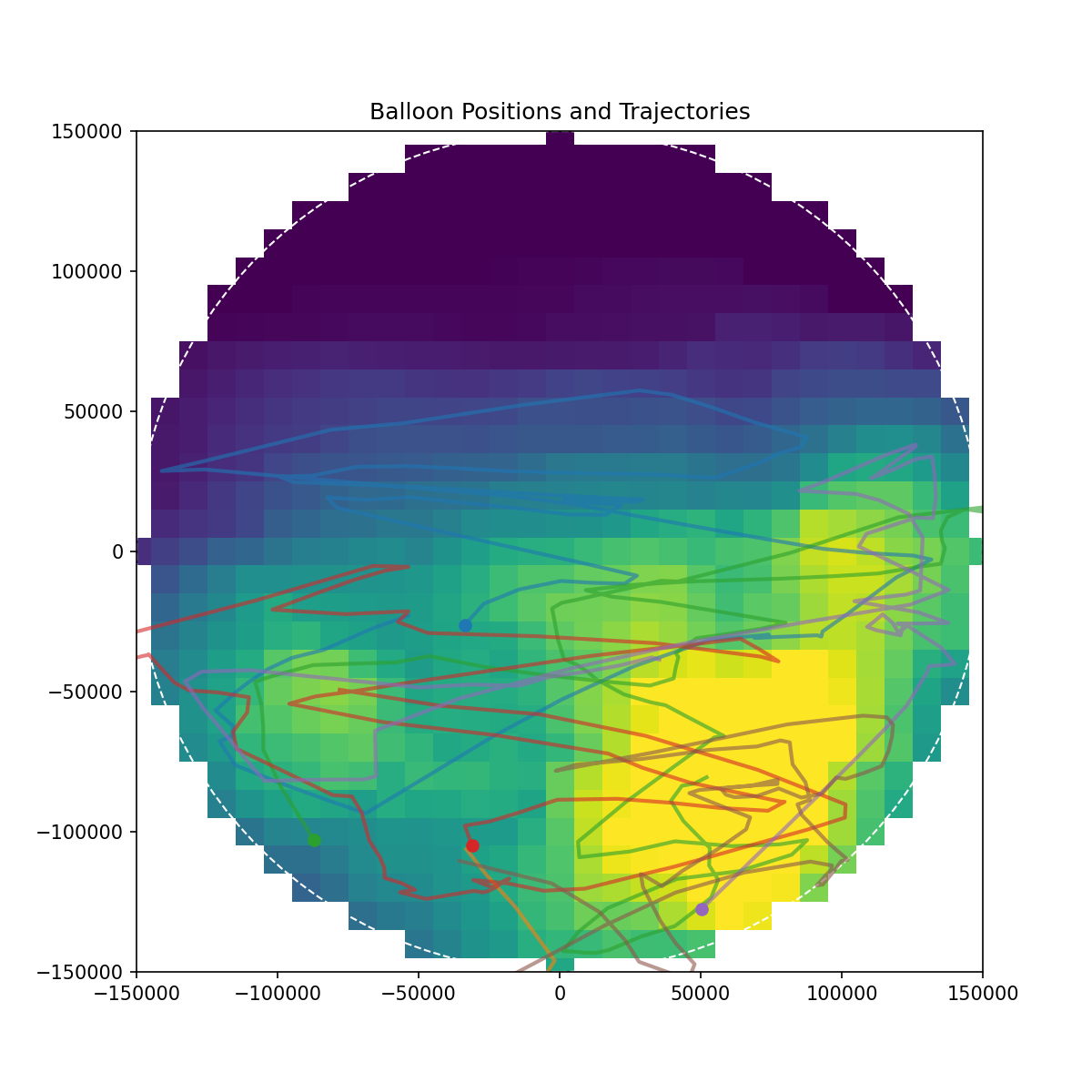}
        \includegraphics[width=0.23\textwidth]{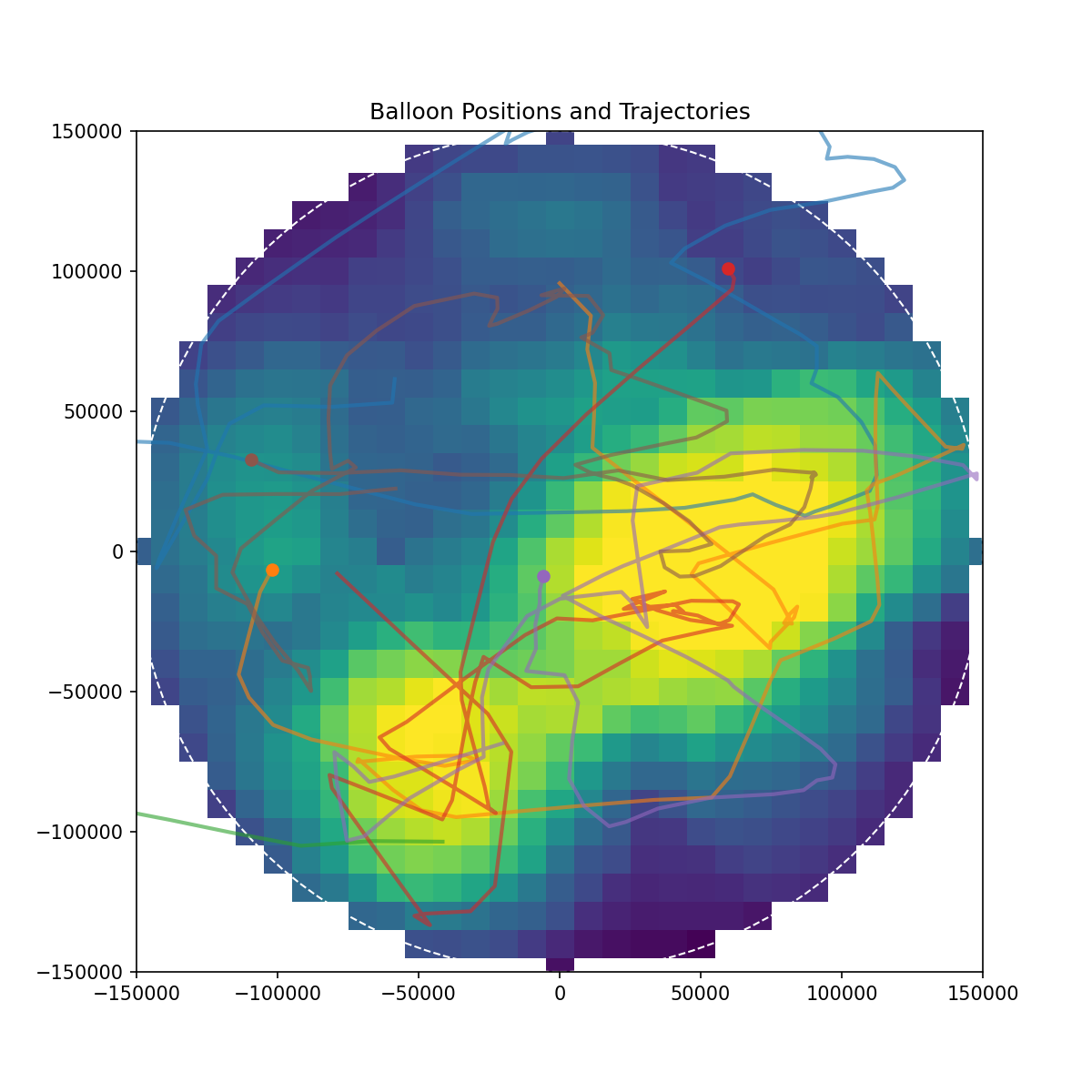}
        \includegraphics[width=0.23\textwidth]{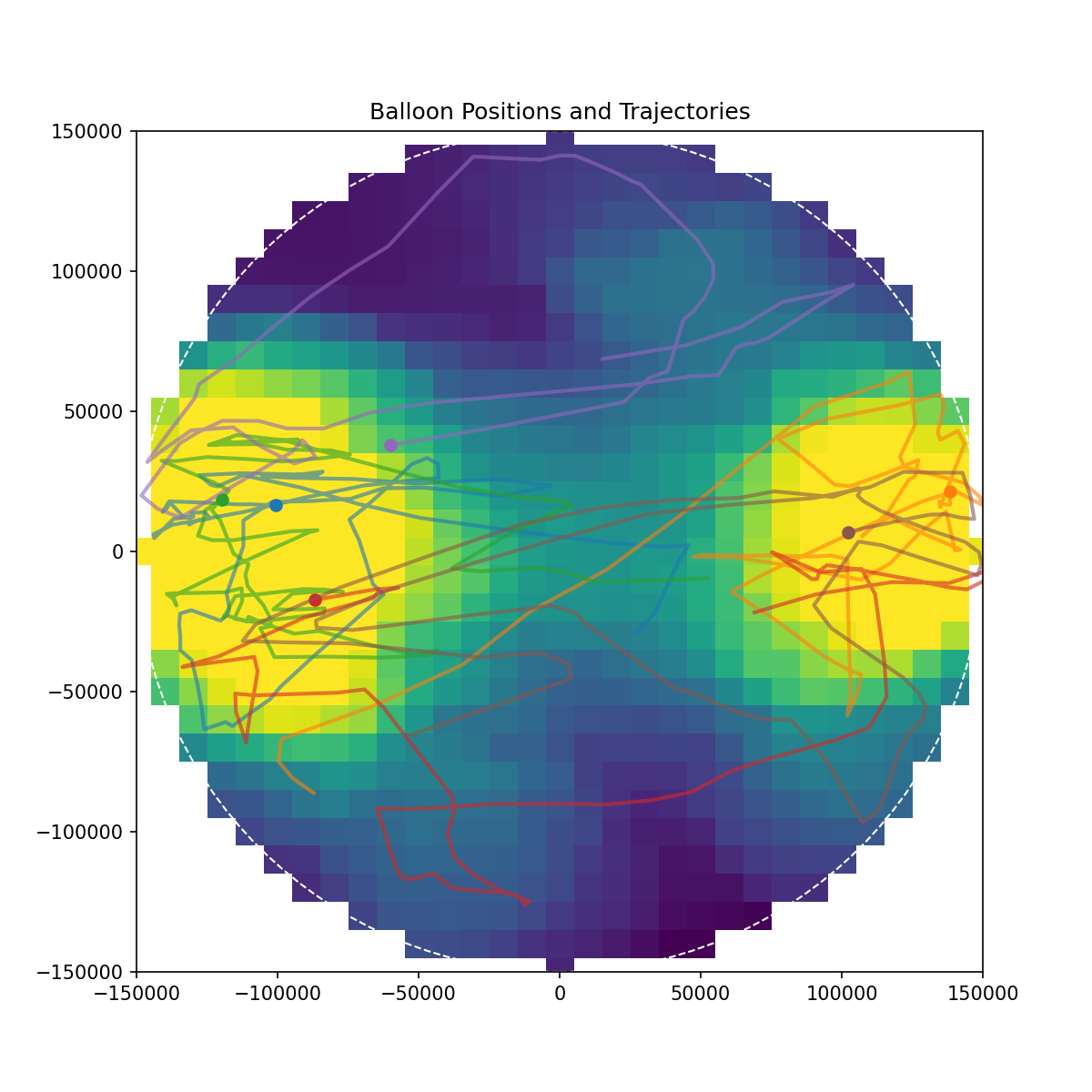}
        \caption{6-Agent QMIX Controller}
        \label{fig:row4}
    \end{subfigure}

    \caption{A sampling of final trajectories and coverage maps between the Voronoi Baseline Controller and QMIX for the same forecast and starting positions. The colored coverage heatmap is capped off at episode length.}
    \label{fig:coverage_trajectories}
\end{figure*}

\textbf{Supplementary Metrics:}
While at the core, both the baseline and QMIX controllers are designed for optimal separation within the TWR, the agents also naturally perform area coverage as a byproduct of this behavior. We perform an initial investigation into evaluating several area coverage metrics.  For simulating area coverage, we generate a coverage heatmap over the duration of an episode, where at each timestep, the HAB has a ground coverage radius of 50 km centered around the HAB's current position.  The maximum value of the heatmap is capped to the total number of episode time steps (2880) so that coverage statistics can be compared directly for different numbers of agents. A sampling of final coverage heatmaps, as well as the individual HAB trajectories for both QMIX and the baseline controller, is shown in Figure~\ref{fig:coverage_trajectories}. 

\subsection{Baseline Performance Analysis}
The Voronoi partitioning baseline demonstrates highly consistent and predictable behavior across all tested configurations. The deterministic nature of the algorithm produces smooth, coordinated trajectories where agents maintain stable formations that gradually translate and rotate as a cohesive unit in response to atmospheric conditions.

Analysis of baseline trajectories reveals several characteristic patterns. Agent waypoints are continuously updated based on Lloyd's relaxation, causing the entire formation to adapt collectively to wind drift. When atmospheric conditions push agents away from their assigned Voronoi centroids, the algorithm smoothly reassigns regions and generates new waypoints that maintain optimal geometric coverage. This results in formations that appear to "breathe" and rotate while preserving spatial relationships between agents. Figure~\ref{fig:coverage_trajectories} subplots (a) and (c) show several final baseline trajectories in various wind conditions and their resulting coverage maps. In ideal HAB-navigable wind conditions, baseline agents typically achieve group station-keeping formations where all agents execute similar altitude maneuvers in unison. Similarly, in poor wind conditions where continuous station-keeping is not possible, the group maintains formation while drifting collectively out of bounds.

%\textbf{Figure: Show specific case with n agents for baseline to show evolution of trajectory over time at 5 different timesteps}

%\textbf{Figure (Appendix): Show few final baseline trajectories for the full range of different number of agents}

\subsection{QMIX Performance Analysis}
QMIX-trained agents exhibit markedly different behavioral patterns characterized by dynamic, adaptive responses to local atmospheric conditions. Unlike the baseline's smooth, predictable movements, QMIX trajectories show more varied and opportunistic navigation strategies that leverage learned coordination policies.

The learned policies demonstrate several emergent behaviors not present in the deterministic baseline. Agents frequently exhibit altitude-switching maneuvers to access favorable wind layers, coordinated repositioning when teammates encounter difficulties, and adaptive coverage reallocation when atmospheric conditions render certain areas inaccessible. These behaviors result in increased trajectory variability, with individual agents taking more diverse paths that exploit local wind patterns and coordinate through shared reward optimization, leading to trajectories that appear more erratic in the short term but demonstrate long-term adaptation to dynamic environmental conditions. Because individual agents in the QMIX controller rarely station keep, coverage hotspots are typically less defined even in favorable HAB-navigable wind conditions, and do not result in well-defined circular regions like with the Baseline, as shown in the final column of Figure~\ref{fig:coverage_trajectories}.

\begin{figure}
\centering
\includegraphics[width=3.4in]{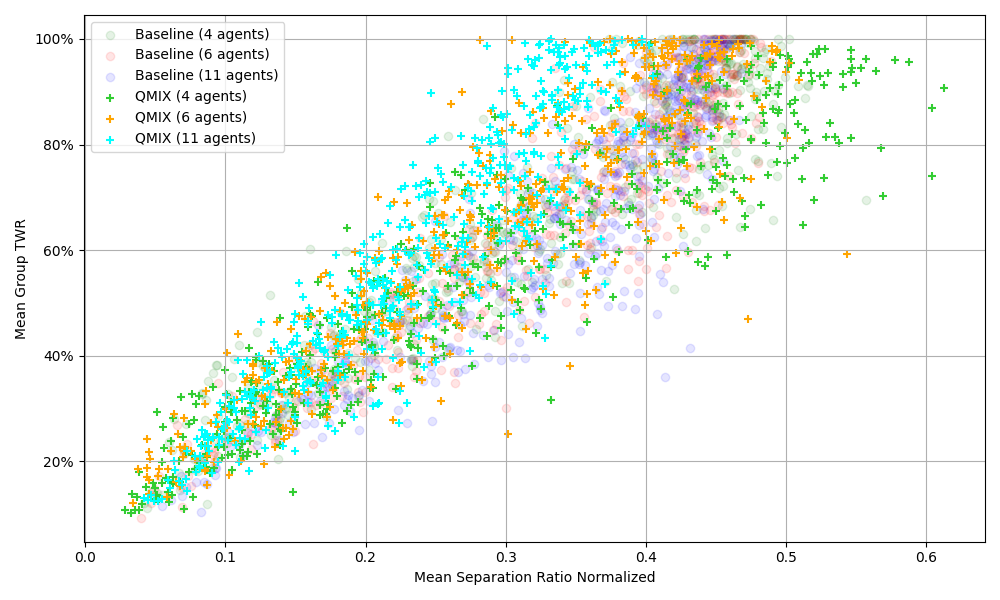}
\caption{Mean Group TWR vs Mean Normalized Separation Ratio between Baseline and QMIX controllers for 4, 6, and 11 agents.}
\label{fig:twr_vs_separation}
\end{figure}

\begin{figure}
\centering
\includegraphics[width=3.4in]{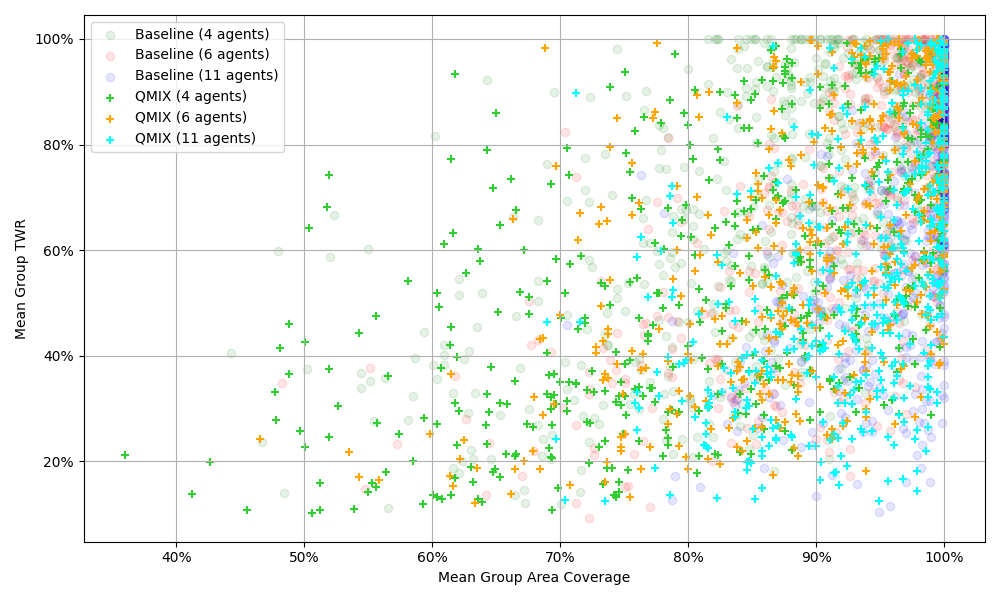}
\caption{Mean Group TWR vs Group Area Coverage between Baseline and QMIX controllers for 4, 6, and 11 agents.}
\label{fig:twr_vs_coverage}
\end{figure}

\subsection{Evaluation Comparison between Baseline vs QMIX Comparisons}
Our two primary evaluation metrics for comparing the baseline controller to the QMIX controller are derived from the 2 rewards used in the QMIX controller, TWR and Separation Ratio.  
The group mean group TWR is equivalent to the mean coverage ratio used in the coverage reward from Eq. \ref{coverage_ratio}. However, the normalized mean separation ratio
uses $d_{target} = D_{coverage}$, instead of $d_{target} = R_{coverage}/\sqrt{N_{agents}}$ from Eq. \ref{dispersion_reward}, so that the  separation metric is comparable between different team sizes.  
Both controllers follow similar distributional trends when comparing these two primary metrics, as shown in Figure~\ref{fig:twr_vs_separation}. As the mean group TWR increases, so does the mean normalized separation ratio between agents. The Baselines controller follows similar separation trends regardless of team size, whereas QMIX has higher separation at hiugh mean group TWRs for 4 agents and less separation than baselines at high group TWRs for 6 agents. 
%Similarly, at higher mean group TWR levels, the mean separation ratio decreases with more agents in the system. 
With the QMIX controller, as the mean TWR approaches 100\%, 4 agents are clustered around a separation ratio of 0.45-0.55, whereas 11 agents are clustered between 0.3-0.4. 
This trend is consistent with the anticipated effect that, in favorable HAB-navigable conditions (where the balloons can continuously stay within the region, TWR approaching 100\%), larger teams will spread out less.

Figure~\ref {fig:twr_vs_coverage} shows that area coverage distributions between Baseline and QMIX are highly varied but also similar. Percent area coverage is taken to be a binary version of the final heatmap (cells covered during the episode or not) in Figure~\ref{fig:coverage_trajectories}, not evaluating for hotspots or persistence.  As more agents are introduced to the system, the distribution clusters more to the top right-hand corner of the scatter plot with 100\% mean group TWR and 100\% mean group coverage.  However, there are still many 6-agent and 11-agent examples from both controllers where lower mean group TWRs ($<$40\%) can still result in approximately full coverage ($>$90\%).  The lower the number of agents, the larger the variability in area coverage performance for both controllers, as shown by the 4-agent distributions.

\begin{figure}
\centering
\includegraphics[width=3.4in]{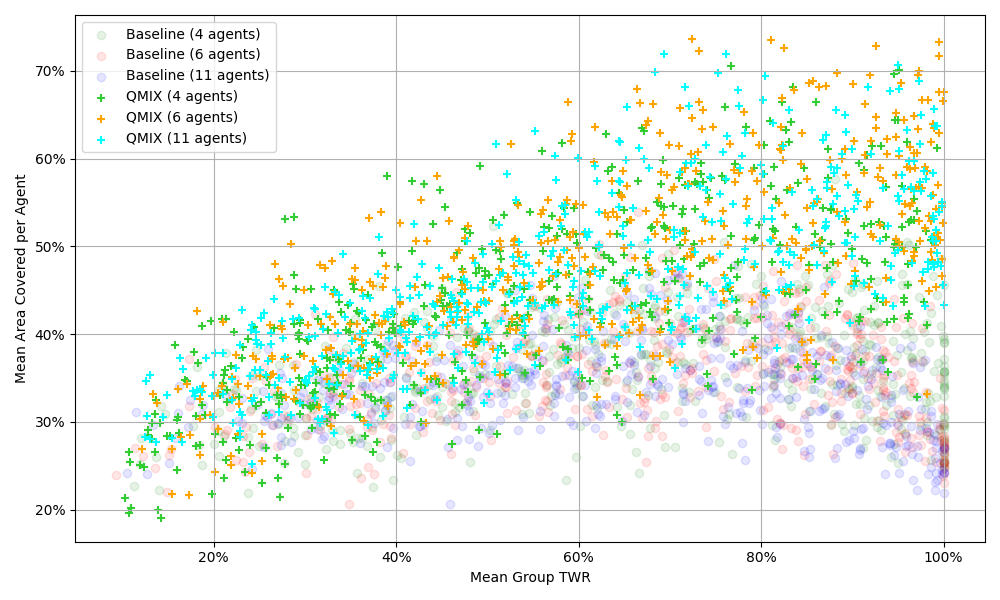}
\caption{Mean Group TWR vs Mean Percent Area Coverage per agent between Baseline and QMIX controllers for 4, 6, and 11 agents.}
\label{fig:group_twr_vs_agent_area}
\end{figure}

\begin{figure}
\centering
\includegraphics[width=3.4in]{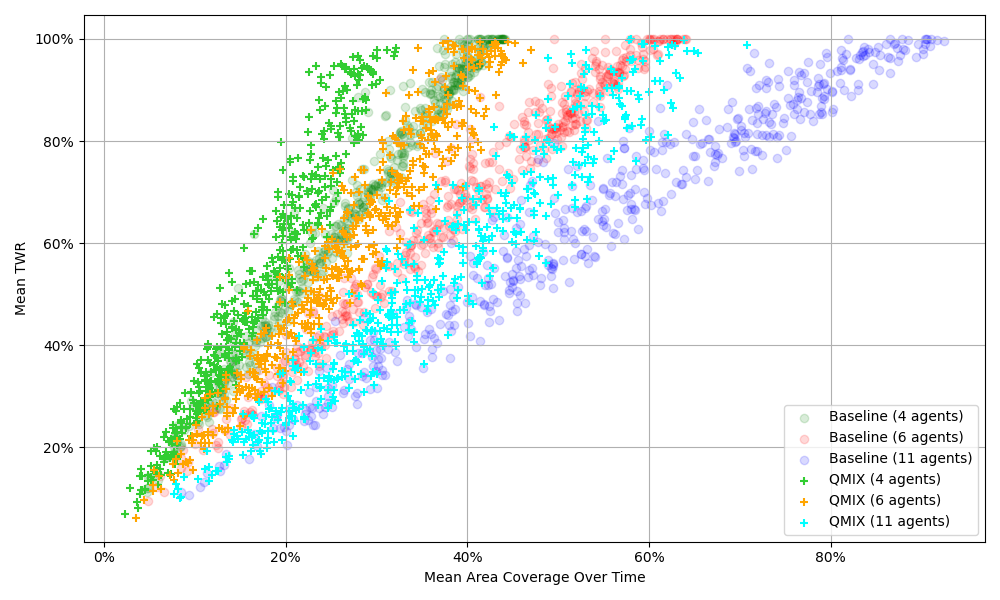}
\caption{Mean Group TWR vs Mean Group Coverage Over Time between Baseline and QMIX controllers for 4, 6, and 11 agents.}
\label{fig:twr_vs_coverage_over_time}
\end{figure}

The largest differences between the baseline and QMIX controllers are how the agents cover the area while attempting to stay within the region.  Figure~\ref{fig:group_twr_vs_agent_area} shows that individual HAB agents in the QMIX controller cover substantially more area when compared with the baseline controller, while maintaining similar mean group TWR numbers. The same plot also shows that at very high mean group TWRs, the baseline controllers' mean area coverage per agent decreases, suggesting station keeping at the ideal waypoints is being achieved and maintained, like in the final column of subplots (a) and (c) in Figure~\ref{fig:coverage_trajectories} showing well-defined hotspots. Figure~\ref{fig:twr_vs_coverage_over_time} shows that overall, both controllers follow 2 similar trends: 1) as the mean group TWR increases, mean group coverage over time also increases, and 2) as more agents are involved, mean group coverage over time increases because more agents within the TWR will naturally cover more area when dispersed.  The baseline controller, however, is better overall, regardless of the number of agents, at maintaining a higher mean group area coverage over time when compared with QMIX. Furthermore, at higher mean group TWRs and numbers of agents, this distributional gap in performance between the baseline and QMIX controller widens. A major contributing factor to the difference in mean coverage over time performance is the fact that the QMIX controller does not penalize agents for operating at the boundaries of the overall target region.  

%In contrast, with the baseline controller at these small to medium team sizes, if the agents could constantly maintain station-keeping at the ideal waypoints, they would achieve optimal coverage over time.

%While overall percent coverage and mean separation distance are typically similar between Baseline and QMIX, there are major differences between the final coverage maps, and the group dynamics and trajectories as shown in Figure \ref{fig:coverage_trajectories}. The baseline controller typically operates as a formation flyer, and QMIX has more random movement.  Unlike the baseline controller, which assigns optimal waypoints and decouples the individual agent control, QMIX controls the actions of all agents without any waypoints. Baseline agents always attempt to station keep at their designated waypoints. In better HAB-navigable wind conditions, the baseline typically results in a group station-keeping formation, taking similar altitude maneuvers in unison. Similarly, in poor wind conditions where continuous station keeping is not possible, the group will typically maintain formation while drifting out of bounds.  This is juxtaposed by the QMIX controller, which is much less predictable, rarely station keeps, and frequently results in individual agents covering much more of the space than the baseline controller in similar conditions. Because individual agents in the QMIX controller do not station keep, the coverage hotspots do not result in well defined circular regions like with the Baseline. 

\subsection{Discussion}

The quantitative results demonstrate that QMIX achieves nearly identical performance to the established Voronoi baseline when comparing metrics included in the QMIX reward function. This result is both expected and significant: for simple distributed coverage with static objectives, near-optimal geometric solutions exist and set the performance ceiling. The fact that QMIX matches these theoretically grounded methods using a relatively simple reward formulation, standard architecture, and no extensive hyperparameter tuning validates that the framework successfully captures the essential coordination dynamics through learned behaviors alone.

The similar performance metrics reveal that both approaches successfully solve the distributed coverage problem, but through fundamentally different operational mechanisms. Voronoi partitioning functions as a global waypoint assignment system that maintains theoretically optimal geometric coverage by continuously computing centroids and reassigning regional responsibilities. In contrast, QMIX operates as a distributed motion control system where individual agents learn altitude control policies that collectively optimize team-level coverage objectives through shared reward optimization.

This distinction becomes critical as mission complexity increases. Geometric optimization methods are inherently limited to problems with closed-form solutions. As mission requirements expand beyond basic coverage to include persistent tracking of dynamic targets, heterogeneous agent capabilities, multi-objective optimization balancing coverage with communication relay and sensor placement, or scenarios requiring predictive adaptation to forecasted atmospheric changes, deterministic approaches become intractable. Learning-based methods, by contrast, can continue to scale with problem complexity. For these more complex missions that extend beyond the simple static coverage task evaluated here, QMIX provides several key advantages. The framework naturally accommodates multi-modal inputs such as global coverage maps, terrain constraints, no-fly zones, and dynamic mission objectives that would require substantial algorithmic modifications to deterministic approaches. The learned policies can potentially respond to variations in operational conditions with limited re-training, particularly valuable when environmental conditions change unpredictably or mission requirements evolve during deployment. Furthermore, QMIX enables heterogeneous agent behavior to emerge naturally from training, allowing different agents to adopt specialized roles within coordination strategies that support complex multi-objective tasks.

Conversely, the baseline Voronoi approach could be enhanced through the incorporation of additional heuristics for more achievable waypoint assignment, predictive atmospheric modeling, and dynamic region prioritization. These improvements to a Voronoi-based approach may also require a different individual HAB agent controller instead of a station-keeping controller. However, such enhancements would increase algorithmic complexity while maintaining the fundamental limitation of predetermined behavioral patterns that cannot adapt to unforeseen mission requirements.

The value of QMIX lies not in superior performance under the current simple distributed coverage task, but in providing a validated learning-based foundation that can accommodate mission complexity growth as HAB applications expand beyond basic area coverage scenarios where geometric optimization remains tractable.

%%%%%%%%%%%%%%%%%%%%%%%%%%%%%%%%%%%%%%
\section{Conclusion and Future Work}
%%%%%%%%%%%%%%%%%%%%%%%%%%%%%%%%%%%%%%

This work presents the first systematic application of multi-agent reinforcement learning for coordinated high-altitude balloon distributed area coverage. We successfully demonstrate that cooperative MARL can match the performance of theoretically optimal geometric methods on the simple task of distributed area coverage, validating the approach and providing a foundation for more complex mission requirements where deterministic methods become intractable.

\subsection{Research Contributions}

Our research makes several key contributions to the intersection of multi-agent reinforcement learning and atmospheric vehicle coordination. We developed the first MARL framework specifically designed for HAB coordination by extending the RLHAB simulation environment to support cooperative multi-agent learning in realistic atmospheric conditions using ERA5 data and synthetic wind fields. This advancement enables systematic study of multi-agent coordination strategies in complex, dynamic atmospheric environments that were previously limited to single-agent or deterministic approaches.

We adapted the QMIX method for high-altitude balloon coordination, demonstrating its effectiveness in learning cooperative policies that match the performance of theoretically grounded geometric optimization methods. This adaptation required a specialized observation space design that addresses the credit assignment problem inherent in shared reward structures, incorporating individual state information, environmental context, and teammate coordination data to enable stable learning under team objectives. Our hierarchical reward structure successfully balances primary coverage maintenance with secondary spatial distribution optimization, ensuring agents prioritize coverage while developing coordinated positioning strategies.

The framework validates that small-team MARL coordination can achieve effective distributed area coverage for practical deployment scenarios, moving beyond large-scale constellation approaches to address realistic HAB mission requirements. Through comprehensive evaluation against the Voronoi partitioning baseline, we demonstrate that learned coordination behaviors can replicate the effectiveness of established geometric distribution methods while offering greater adaptability potential for complex mission requirements that extend beyond basic area coverage optimization.

\subsection{Limitations}

QMIX exhibits significant scalability limitations that constrain its applicability to larger HAB fleets or constellations. The observation space dimensionality grows as $6 + 3 \times N_{levels} + 5 \times (N_{agents} - 1)$, creating computational bottlenecks as team size increases beyond 5-7 agents. For larger configurations, the high-dimensional observation space approaches the practical limits of simple network architectures, requiring extended training periods.

The global state space scaling as $N_{agents} \times (6 + 3 \times N_{levels}) + 4$ presents additional challenges, with dimensional explosion necessitating careful hyperparameter tuning. Unlike the Voronoi baseline, which adapts automatically to any team configuration, QMIX requires complete retraining for different agent counts, limiting operational flexibility for missions with variable or changing fleet sizes.  In small to medium areas in predictable, trendy wind environments, the forecast observation space may be able to be reduced to a global estimate, rather than having a wind column observation for each agent.

Memory requirements scale correspondingly, with replay buffer storage and network parameter counts creating practical deployment constraints. The current framework assumes circular coverage areas and would require architectural modifications for irregular or non-convex regions. These limitations suggest our current formulation of QMIX is most suitable for small-to-medium HAB teams rather than large-scale constellations.

\subsection{Future Work}

Several future research directions emerge from this foundational work in HAB MARL coordination. Developing more scalable MARL architectures that maintain coordination effectiveness while reducing computational complexity could enable applications to larger fleets. Investigation of curriculum learning approaches has the potential to accelerate training convergence and improve sample efficiency for complex atmospheric conditions. Extension to heterogeneous agent capabilities, irregular coverage geometries, and multi-objective mission requirements is another important advancement opportunity. Additionally, the integration of real-time atmospheric forecasting uncertainty and adaptive mission planning could enhance practical deployment capabilities. Exploring transfer learning approaches to enable policies trained in one geographic region or season to generalize to new conditions could significantly reduce deployment costs and training requirements. Finally, validation through real-world HAB flight tests would provide crucial insights into the transition from simulation to real operational environments.

%The framework established here provides a foundation for exploring these advanced coordination challenges while maintaining the practical focus on deployable atmospheric vehicle systems that can adapt to the evolving complexity of HAB mission requirements.

%%%%%%%%%%%%%%%%%%%%%%%%%%%%%%%%%%%%%%%%%%%%%%%%%%%%%%%%%%%%%%%%%%%%%%%%%%%%%%%%%%%%%%%%%%%%%%%%%%%%%%
\acknowledgements
%This research was conducted in part during an NREIP internship at the U.S. Naval Research Laboratory. 
The authors acknowledge the support of Iowa State University for providing additional computational resources and research facilities.

%%%%%%%%%%%%%%%%%%%%%%%%%%%%%%%%%%%%%%%%%%%%%%%%%%%%%%%%%%%%%%%%%%%%%%%%%%%%%%%%%%%%%%%%%%%%%%%%%%%%%%
\bibliographystyle{IEEEtran}
\bibliography{references}

\begin{thebibliography}{10}
\providecommand{\url}[1]{#1}
\csname url@samestyle\endcsname
\providecommand{\newblock}{\relax}
\providecommand{\bibinfo}[2]{#2}
\providecommand{\BIBentrySTDinterwordspacing}{\spaceskip=0pt\relax}
\providecommand{\BIBentryALTinterwordstretchfactor}{4}
\providecommand{\BIBentryALTinterwordspacing}{\spaceskip=\fontdimen2\font plus
\BIBentryALTinterwordstretchfactor\fontdimen3\font minus \fontdimen4\font\relax}
\providecommand{\BIBforeignlanguage}[2]{{%
\expandafter\ifx\csname l@#1\endcsname\relax
\typeout{** WARNING: IEEEtran.bst: No hyphenation pattern has been}%
\typeout{** loaded for the language `#1'. Using the pattern for}%
\typeout{** the default language instead.}%
\else
\language=\csname l@#1\endcsname
\fi
#2}}
\providecommand{\BIBdecl}{\relax}
\BIBdecl

\bibitem{abe2009scientific}
T.~Abe, T.~Imamura, N.~Izutsu, and N.~Yajima, \emph{Scientific Ballooning: Technology and Applications of Exploration Balloons Floating in the Stratosphere and the Atmospheres of Other Planets}.\hskip 1em plus 0.5em minus 0.4em\relax Springer, 2009.

\bibitem{schuler2025winddiversity}
T.~K. Schuler and C.~Motell, ``Wind diversity trends in the lower stratosphere analyzed from radiosondes launched in the western hemisphere,'' \emph{Journal of Geophysical Research: Atmospheres}, vol. 130, no.~10, p. e2024JD042770, 2025.

\bibitem{brown2025stratospheric}
D.~Brown and J.~Leidich, ``The stratospheric goldilocks zone is critical for high-altitude balloon navigation,'' \emph{Communications Earth \& Environment}, vol.~6, no.~1, p. 544, 2025.

\bibitem{koch2019reinforcement}
W.~Koch, R.~Mancuso, R.~West, and A.~Bestavros, ``Reinforcement learning for uav attitude control,'' \emph{ACM Transactions on Cyber-Physical Systems}, vol.~3, no.~2, pp. 1--21, 2019.

\bibitem{patino2023learning}
D.~Pati{\~n}o, S.~Mayya, J.~Calderon, K.~Daniilidis, and D.~Salda{\~n}a, ``Learning to navigate in turbulent flows with aerial robot swarms: A cooperative deep reinforcement learning approach,'' \emph{IEEE Robotics and Automation Letters}, vol.~8, no.~7, pp. 4219--4226, 2023.

\bibitem{cai2024reinforcement}
Y.~Cai, K.-S. Low, and Z.~Wang, ``Reinforcement learning-based satellite formation attitude control under multi-constraint,'' \emph{Advances in Space Research}, vol.~74, no.~11, pp. 5819--5836, 2024.

\bibitem{bellemare2020autonomous}
M.~G. Bellemare, S.~Candido, P.~S. Castro, J.~Gong, M.~C. Machado, S.~Moitra, S.~S. Ponda, and Z.~Wang, ``Autonomous navigation of stratospheric balloons using reinforcement learning,'' \emph{Nature}, vol. 588, no. 7836, pp. 77--82, 2020.

\bibitem{gannetti2023navigation}
M.~Gannetti, M.~Gemignani, and S.~Marcuccio, ``Navigation of sounding balloons with deep reinforcement learning,'' in \emph{2023 IEEE 10th International Workshop on Metrology for AeroSpace (MetroAeroSpace)}.\hskip 1em plus 0.5em minus 0.4em\relax IEEE, 2023, pp. 591--596.

\bibitem{schuler2025seasonal}
T.~K. Schuler, C.~Prasad, G.~Kiselev, and D.~Sofge, ``Seasonal station-keeping of short duration high altitude balloons using deep reinforcement learning,'' \emph{arXiv preprint arXiv:2502.05014}, 2025.

\bibitem{xing2024multi}
X.~Xing, Z.~Zhou, Y.~Li, B.~Xiao, and Y.~Xun, ``Multi-uav adaptive cooperative formation trajectory planning based on an improved matd3 algorithm of deep reinforcement learning,'' \emph{IEEE Transactions on Vehicular Technology}, vol.~73, no.~9, pp. 12\,484--12\,499, 2024.

\bibitem{yun2022cooperative}
W.~J. Yun, S.~Park, J.~Kim, M.~Shin, S.~Jung, D.~A. Mohaisen, and J.-H. Kim, ``Cooperative multiagent deep reinforcement learning for reliable surveillance via autonomous multi-uav control,'' \emph{IEEE Transactions on Industrial Informatics}, vol.~18, no.~10, pp. 7086--7096, 2022.

\bibitem{tang2025enhanced}
R.~Tang, J.~Tang, M.~S.~A. Talip, N.~K. Aridas, and X.~Xu, ``Enhanced multi agent coordination algorithm for drone swarm patrolling in durian orchards,'' \emph{Scientific Reports}, vol.~15, no.~1, p. 9139, 2025.

\bibitem{rashid2020monotonic}
T.~Rashid, M.~Samvelyan, C.~S. De~Witt, G.~Farquhar, J.~Foerster, and S.~Whiteson, ``Monotonic value function factorisation for deep multi-agent reinforcement learning,'' \emph{Journal of Machine Learning Research}, vol.~21, no. 178, pp. 1--51, 2020.

\bibitem{du2022dynamic}
H.~Du, T.~Sun, M.~Lv, M.~Junhui, and Z.~Zhang, ``Dynamic coverage performance of wind-assisted balloons mesh based on voronoi partition and energy constraint,'' \emph{Advances in Space Research}, vol.~70, no.~2, pp. 470--484, 2022.

\bibitem{vandermeulen2017distributed}
I.~Vandermeulen, M.~Guay, and P.~J. McLellan, ``Distributed control of high-altitude balloon formation by extremum-seeking control,'' \emph{IEEE Transactions on Control Systems Technology}, vol.~26, no.~3, pp. 857--873, 2017.

\bibitem{xu2022station}
Z.~Xu, Y.~Liu, H.~Du, and M.~Lv, ``Station-keeping for high-altitude balloon with reinforcement learning,'' \emph{Advances in Space Research}, vol.~70, no.~3, pp. 733--751, 2022.

\bibitem{huh2023multi}
D.~Huh and P.~Mohapatra, ``Multi-agent reinforcement learning: A comprehensive survey,'' \emph{arXiv preprint arXiv:2312.10256}, 2023.

\bibitem{hernandez2017survey}
P.~Hernandez-Leal, M.~Kaisers, T.~Baarslag, and E.~M. De~Cote, ``A survey of learning in multiagent environments: Dealing with non-stationarity,'' \emph{arXiv preprint arXiv:1707.09183}, 2017.

\bibitem{gronauer2022multi}
S.~Gronauer and K.~Diepold, ``Multi-agent deep reinforcement learning: a survey,'' \emph{Artificial Intelligence Review}, vol.~55, no.~2, pp. 895--943, 2022.

\bibitem{zhang2021multi}
K.~Zhang, Z.~Yang, and T.~Ba{\c{s}}ar, ``Multi-agent reinforcement learning: A selective overview of theories and algorithms,'' \emph{Handbook of reinforcement learning and control}, pp. 321--384, 2021.

\bibitem{tan1993multi}
M.~Tan, ``Multi-agent reinforcement learning: Independent vs. cooperative agents,'' in \emph{Proceedings of the tenth international conference on machine learning}, 1993, pp. 330--337.

\bibitem{littman1994markov}
M.~L. Littman, ``Markov games as a framework for multi-agent reinforcement learning,'' in \emph{Machine learning proceedings 1994}.\hskip 1em plus 0.5em minus 0.4em\relax Elsevier, 1994, pp. 157--163.

\bibitem{yang2020overview}
Y.~Yang and J.~Wang, ``An overview of multi-agent reinforcement learning from game theoretical perspective,'' \emph{arXiv preprint arXiv:2011.00583}, 2020.

\bibitem{yuan2023survey}
L.~Yuan, Z.~Zhang, L.~Li, C.~Guan, and Y.~Yu, ``A survey of progress on cooperative multi-agent reinforcement learning in open environment,'' \emph{arXiv preprint arXiv:2312.01058}, 2023.

\bibitem{sunehag2017value}
P.~Sunehag, G.~Lever, A.~Gruslys, W.~M. Czarnecki, V.~Zambaldi, M.~Jaderberg, M.~Lanctot, N.~Sonnerat, J.~Z. Leibo, K.~Tuyls \emph{et~al.}, ``Value-decomposition networks for cooperative multi-agent learning,'' \emph{arXiv preprint arXiv:1706.05296}, 2017.

\bibitem{lowe2017multi}
R.~Lowe, Y.~I. Wu, A.~Tamar, J.~Harb, O.~Pieter~Abbeel, and I.~Mordatch, ``Multi-agent actor-critic for mixed cooperative-competitive environments,'' \emph{Advances in neural information processing systems}, vol.~30, 2017.

\bibitem{foerster2018counterfactual}
J.~Foerster, G.~Farquhar, T.~Afouras, N.~Nardelli, and S.~Whiteson, ``Counterfactual multi-agent policy gradients,'' in \emph{Proceedings of the AAAI conference on artificial intelligence}, vol.~32, no.~1, 2018.

\bibitem{amato2024introduction}
C.~Amato, ``An introduction to centralized training for decentralized execution in cooperative multi-agent reinforcement learning,'' \emph{arXiv preprint arXiv:2409.03052}, 2024.

\end{thebibliography}
%%%%%%%%%%%%%%%%%%%%%%%%%%%%%%%%%%%%%%%%%%%%%%%%%%%%%%%%%%%%%%%%%%%%%%%%%%%%%%%%%%%%%%%%%%%%%%%%%%%%%%

%%%%%%%%%%%%%%%%%%%%%%%%%%%%%%%%%%%%%%%%%%%%%%%%%%%%%%%%%%%%%%%%%%%%%%%%%%%%%%%%%%%%%%%%%%%%%%%%%%%%%%
\thebiography
%% This biostyle allows you to insert your photo size 1in X 1.25in
\begin{biographywithpic}
{Adam Haroon}{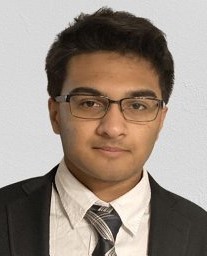}
is an undergraduate robotics researcher at Iowa State University in the Coordinated Systems Laboratory, pursuing degrees in Applied Mathematics and Computer Science. His work focuses on autonomous robotic perception, deep reinforcement learning, and multi-agent systems, with applications in urban air mobility and robot navigation. Adam has previously worked at the U.S. Naval Research Laboratory’s Autonomous Systems Laboratory as an NREIP intern and at the U.S. Department of Defense DEVCOM Soldier Center as an NSIN X-Force Fellow, contributing to 4D human body reconstruction for PPE evaluation and robotic learning. His work has been published in SPIE and IEEE conferences, and his research has been recognized with the Boeing Undergraduate Research Fellowship, Navy Engineering Analytics Program Scholarship, and NASA Undergraduate Research Scholarship.
\end{biographywithpic} 

\begin{biographywithpic}
{Tristan Schuler}{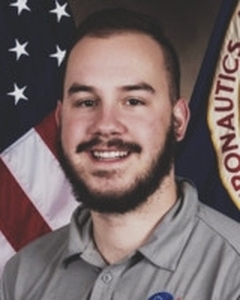}
is a Robotics Research Scientist at the U.S. Naval Research Laboratory (NRL) in the Distributed Autonomous Systems Section for the Navy Center for Applied Research in Artificial Intelligence (NCARAI). One of his primary research focus areas is in autonomous lighter-than-air platforms with many publications and patents since 2018. Tristan has a B.S. in Mechanical Engineering from George Mason University and an M.S. in Aerospace Engineering from the University of Arizona where he defended his thesis on using solar balloons for planetary exploration. In 2021, he was awarded a Karles Research Fellowship from the NRL to develop autonomous HAB technology for SHAB-Vs.
\end{biographywithpic}

\end{document}